\documentclass[lettersize,journal]{IEEEtran}
\usepackage{amsmath,amsfonts}
\usepackage{algorithmic}
\usepackage{algorithm}
\usepackage{array}
\usepackage[caption=false,font=normalsize,labelfont=sf,textfont=sf]{subfig}
\usepackage{textcomp}
\usepackage{stfloats}
\usepackage{url}
\usepackage{verbatim}
\usepackage{graphicx}
\usepackage{cite}
\usepackage{booktabs}   
\usepackage{multirow}   
\usepackage{array}      
\usepackage{makecell}
\usepackage[table]{xcolor}
\usepackage{adjustbox} 

\newcommand{\pos}[1]{\textcolor{blue!80!black}{#1}}   
\newcommand{\negv}[1]{\textcolor{green!80!black}{#1}} 

\begin{document}

\title{LongFly: Long-Horizon UAV Vision-and-Language Navigation \\ with Spatiotemporal Context Integration}


\author{Wen Jiang, Li Wang, Kangyao Huang, Wei Fan, Jinyuan Liu, Shaoyu Liu,\\ Hongwei Duan, Bin Xu, and Xiangyang Ji, ~\IEEEmembership{Member,~IEEE }%

\thanks{
This work was supported by the National Natural Science Foundation of China under Grant No. 52502496, U22B2052 and the Natural Science Foundation of Chongqing, China under Grant No. CSTB2025NSCQ-GPX0413, and the National High Technology Research and Development Program of China under Grant No. 2020YFC1512501. (Corresponding authors: Bin Xu and Xiangyang Ji)}

\thanks{Wen Jiang, Li Wang, Wei Fan, Hongwei Duan, and Bin Xu are with the School of Mechanical Engineering, Beijing Institute of Technology, Beijing 100081, China. (e-mail: 3120235086@bit.edu.cn, wangli\_bit@bit.edu.cn,fanweixx@bit.edu.cn, 3220250437@bit.edu.cn, bitxubin@bit.edu.cn, }
\thanks{Li Wang is also with the Chongqing Innovation Center, Beijing Institute of Technology, Chongqing 401120, China.}
\thanks{Kangyao Huang is with the Department of Computer Science and Technology, Tsinghua University, Beijing 100084, China.(e-mail: huangky22@mails.tsinghua.edu.cn)}
\thanks{Xiangyang Ji is with the Department of Automation, Tsinghua University, Beijing 100084, China.(e-mail: xyji@tsinghua.edu.cn)}
\thanks{Jinyuan Liu is with the School of Software, Dalian University of Technology, Dalian 116024, China.(e-mail: jinyuanliu@dlut.edu.cn)}
\thanks{Shaoyu Liu is with the School of Artificial Intelligence, Xidian University, Xi'an 710071, China.(e-mail: 23171110721@stu.xidian.edu.cn)}

}

\markboth{Journal of \LaTeX\ Class Files,~Vol.~14, No.~8, August~2021}%
{Shell \MakeLowercase{\textit{et al.}}: A Sample Article Using IEEEtran.cls for IEEE Journals}

\IEEEpubid{0000--0000/00\$00.00~\copyright~2021 IEEE}

\maketitle

\begin{abstract}
Unmanned aerial vehicles (UAVs) are crucial tools for post-disaster search and rescue, facing challenges such as high information density, rapid changes in viewpoint, and dynamic structures, especially in long-horizon navigation. However, current UAV vision-and-language navigation(VLN) methods struggle to model long-horizon spatiotemporal context in complex environments, resulting in inaccurate semantic alignment and unstable path planning. To this end, we propose LongFly, a spatiotemporal context modeling framework for long-horizon UAV VLN. LongFly proposes a history-aware spatiotemporal modeling strategy that transforms fragmented and redundant historical data into structured, compact, and expressive representations.
First, we propose the slot-based historical image compression module, which dynamically distills multi-view historical observations into fixed-length contextual representations. Then, the spatiotemporal trajectory encoding module is introduced to capture the temporal dynamics and spatial structure of UAV trajectories. Finally, to integrate existing spatiotemporal context with current observations, we design the prompt-guided multimodal integration module to support time-based reasoning and robust waypoint prediction. Experimental results demonstrate that LongFly outperforms state-of-the-art UAV VLN baselines by 7.89\% in success rate and 6.33\% in success weighted by path length, consistently across both seen and unseen environments.
\end{abstract}

\begin{IEEEkeywords}
Long-horizon navigation, multimodal prompt fusion, spatiotemporal context modeling, unmanned aerial vehicle, vision-and-language navigation.
\end{IEEEkeywords}

\section{Introduction}
\label{sec:introduction}

\IEEEPARstart{T}{HE} demand for rapid and efficient geospatial data collection~\cite{10048552} and environmental monitoring~\cite{10570242} is becoming increasingly urgent, especially for remote sensing tasks~\cite{10640736} in complex terrains and large areas~\cite{fang2025taskorientedcommunicationsvisualnavigation}. UAVs\cite{11133703,10476501},\cite{zhang2025coordfieldcoordinationfieldagentic} have become the primary choice to meet this need due to their autonomy and efficiency in these tasks. However, traditional unmanned aerial vehicles (UAVs) navigation systems often perform poorly in complex environments due to a lack of semantic reasoning and task understanding, limiting their ability to interpret high-level objectives. With the rapid development of vision-and-language navigation (VLN) technology\cite{pokhrel2025harnessing,anderson2018vision},~\cite{wang2023gridmmgridmemorymap},~\cite{Zheng_2024,HE2024110511}, navigation has been redefined as a task of understanding and executing natural language instructions through environmental visual information. This paradigm enables agents to interpret semantic goals, perceive spatial layouts, and dynamically adjust paths according to task requirements. By aligning visual perception with natural language understanding, UAVs have significantly enhanced their autonomy in complex and GPS-denied environments~\cite{10356107}. In recent years, UAV VLN~\cite{xiao2025uavonbenchmarkopenworldobject,guo2025bedicomprehensivebenchmarkevaluating} has shown strong potential and capabilities in unknown environment navigation, semantic scene understanding, and high-level task execution. While the success of UAV VLN in short-range navigation and atomic instruction execution, the long-horizon dependency challenge for UAV VLN remain largely unexplored.
\begin{figure*}
    \centering
    \includegraphics[width=1\linewidth]{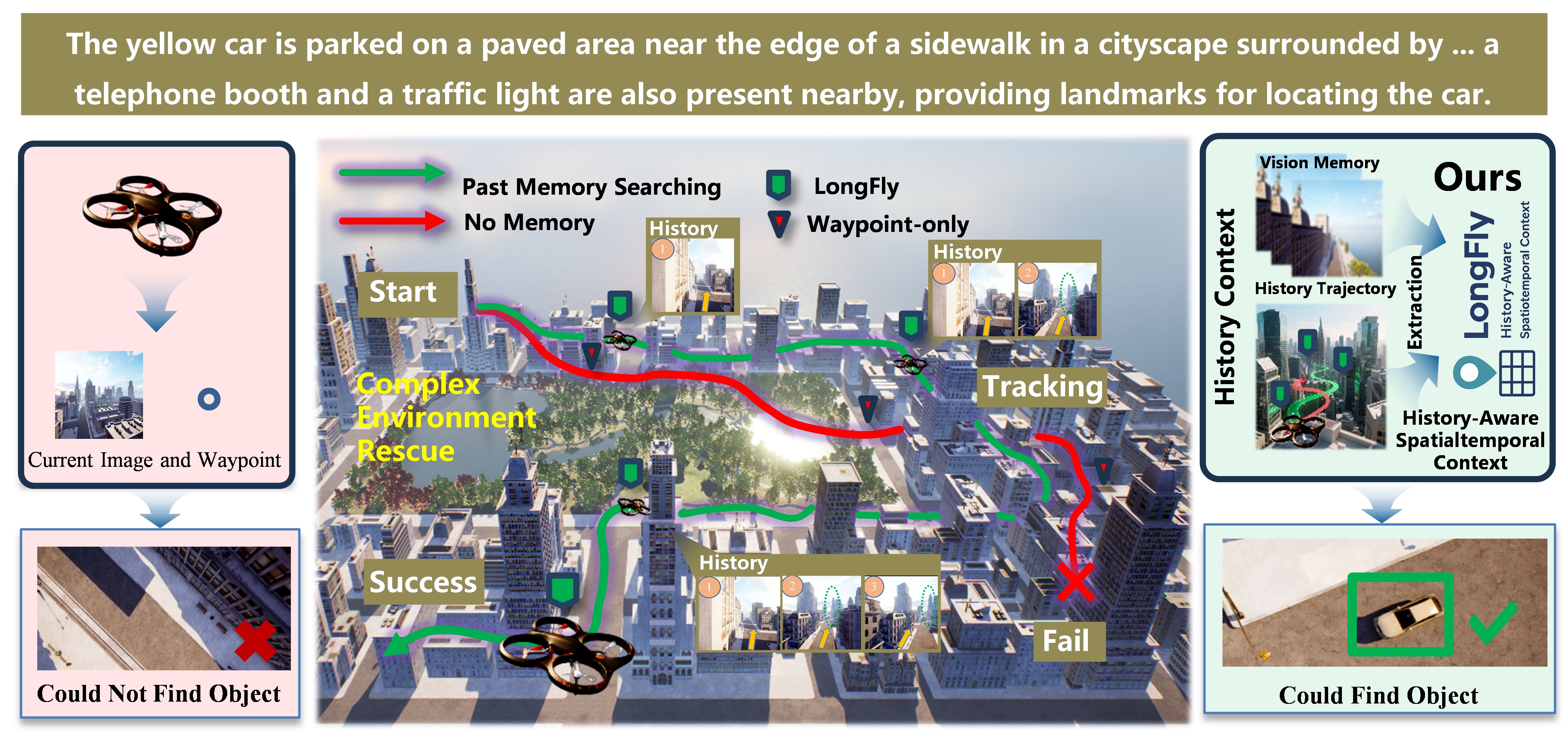}
    \caption{Effect of spatiotemporal context integration on UAV VLN.
    Left (red): Navigation based only on the current image and waypoint fails under rapid viewpoint and layout changes.
    Right (green): LongFly introduces a spatiotemporal context modeling framework for long-horizon UAV VLN. LongFly dynamically distills multi-view historical observations into a compact and semantically rich representation, encodes UAV trajectory dynamics, and aligns spatiotemporal context with current observations and language instructions for robust navigation in complex 3D environments.}
    \label{fig:enter-label}
\end{figure*}

\IEEEpubidadjcol  

Most existing UAV VLN methods struggle with long-horizon dependency problem because they lack a unified spatiotemporal context, which hurts navigation stability and accuracy.
Pioneering works such as AerialVLN~\cite{liu2023aerialvln} and AVDN~\cite{fan2023aerial} are the first to propose that UAVs navigate based on natural language, successfully executing short-range tasks through atomic instructions. To tackle long-horizon dependen~\cite{pardyl2025flysearchexploringvisionlanguagemodels},
Subsequent efforts, such as Citynav~\cite{lee2024citynav} and TravelUAV\cite{wang2024towards}, incorporate large models to better understand the more complex language and environments involved in long-horizon dependencies~\cite{serpiva2025racevla,cai2025flightgpt},\cite{sautenkov2025uav,zhang2025grounded}. Recent efforts explore long-horizon UAV navigation through reinforcement learning~\cite{lin2025openvln,LV2025110075},\cite{cai2025sagcssemanticawaregaussiancurriculum}, BEV-based representations~\cite{zhao2025aerialvisionandlanguagenavigationgridbased}, and various mapping~\cite{ji2025autonomousuavvisualobject} and memory mechanisms~\cite{zhang2025citynavagentaerialvisionandlanguagenavigation,gao2025openfly}. These approaches aim to enhance long-horizon planning and history utilization, but their effectiveness remains limited in complex long-horizon scenarios.
While existing models show impressive performance in short-range navigation, their effectiveness degrades significantly in long-horizon tasks. Although various memory and history mechanisms have been explored, historical information is still mostly treated as static cues and remains weakly connected to the spatiotemporal structure of the navigation process.

Long-horizon dependency remains a major challenge in UAV VLN~\cite{zhang2025groundedvisionlanguagenavigationuavs,song2025longhorizonvisionlanguagenavigationplatform}. 
In addressing this challenge, the following two issues are critical: (1) Adaptive selection and active retrieval of historical spatiotemporal information.
Long-horizon flights continuously accumulate large amounts of highly redundant visual observations and trajectory data. How to adaptively extract the information that is most relevant to the current language instruction and navigation decision from an ever-growing sequence, while effectively suppressing redundant and noisy spatiotemporal cues, remains a key challenge for improving long-horizon decision stability. (2) Effective alignment and integration of multimodal spatiotemporal context. Visual observations, flight trajectories, and language instructions are inherently disjointed. Simple feature stacking fails to capture the logical links between "past movements" and "current goals." Without this alignment, UAVs easily lose their sense of spatial context, leading to inconsistent navigation behaviors over long distances.

To address the challenges discussed above, we propose LongFly, a spatiotemporal context modeling framework as shown in Fig.~\ref{fig:enter-label}. LongFly proposes a history-aware spatiotemporal modeling strategy that transforms fragmented and redundant historical data into structured, compact, and expressive representations. Specifically, we propose the slot-based historical image compression module, which dynamically distills multi-view historical observations into fixed-length contextual representations. Then, the spatiotemporal trajectory encoding module is introduced to capture the temporal dynamics and spatial structure of UAV trajectories. Finally, to integrate existing spatiotemporal context with current observations, we design the prompt-guided multimodal integration module to support time-based reasoning and robust waypoint prediction.

Our main contributions can be summarized as follows:

\begin{itemize}
    \item we propose LongFly, a spatiotemporal context modeling framework for long-horizon UAV VLN, ensuring consistent global decision making in complex 3D environments. 
    \item We introduce a history aware modeling strategy that transforms fragmented historical observations into a structured representation. This approach distills salient information and captures trajectory evolution to support stable long-horizon navigation. 
    \item Experimental results demonstrate that LongFly outperforms state-of-the-art UAV VLN baselines by 7.89\% in success rate and 6.33\% in success weighted by path length, consistently across both seen and unseen environments.
\end{itemize}

The remainder of this paper is structured as follows. Section~\ref{sec:related_work} first surveys vision-and-language navigation for UAVs and long-horizon vision-and-language navigation for UAVs. Section~\ref{sec:method} introduces the proposed LongFly framework, detailing the UAV VLN task formulation and module design. Section~\ref{sec:experiment_setup} describes the experimental setup, including datasets, implementation details, and evaluation metrics. Section~\ref{sec:experimental_results} presents experimental results and analyses to evaluate the effectiveness of LongFly. Finally, Section~\ref{sec:conclusions} summarizes our findings on the effectiveness of explicit spatiotemporal modeling in long-horizon UAV VLN.

\section{RELATED WORK}

\label{sec:related_work}

\subsection{Vision-and-language navigation for UAVs}
 VLN tasks have gradually expanded from indoor to outdoor dynamic environments~\cite{xiang2025navr2dualrelationreasoninggeneralizable}. For example, AerialVLN~\cite{liu2023aerialvln} and AVDN~\cite{fan2023aerial} are the first to propose that UAVs navigate by natural language instructions~\cite{hu2025seepointflylearningfree}. With the integration of large language models\cite{10948476}, methods such as CityNav~\cite{lee2024citynav} and OpenFly~\cite{gao2025openfly} combine models like GPT~\cite{brown2020languagemodelsfewshotlearners}, Vicuna~\cite{zheng2023judgingllmasajudgemtbenchchatbot} and LLaVA~\cite{liu2023visualinstructiontuning} with visual inputs, allowing UAVs to accomplish tasks such as object recognition, obstacle avoidance, and path planning, while achieving improvements in generalization and human-like planning behaviors~\cite{chen2025gradnavvisionlanguagemodelenabled,verraest2025skydreamerinterpretableendtoendvisionbased}. VLA-AN~\cite{wu2025vlaanefficientonboardvisionlanguageaction} proposes a vision–language framework for UAV navigation that integrates visual perception and language understanding to guide autonomous flight. Meanwhile, the TravelUAV project~\cite{wang2024towards} established a UAV VLN benchmark in the AirSim simulator~\cite{xiao2025uavonbenchmarkopenworldobject}, further advancing research in this area. Such developments mark a significant step in UAV VLN evolution, moving from constrained indoor tasks to realistic outdoor ~\cite{ferrag2025uavbenchopenbenchmarkdataset}. Overall, these advances set the stage for tackling new challenges like long-horizon aerial navigation, which we explore in the following section.

\begin{figure*}[!t]
    \centering
    \includegraphics[width=1\textwidth]{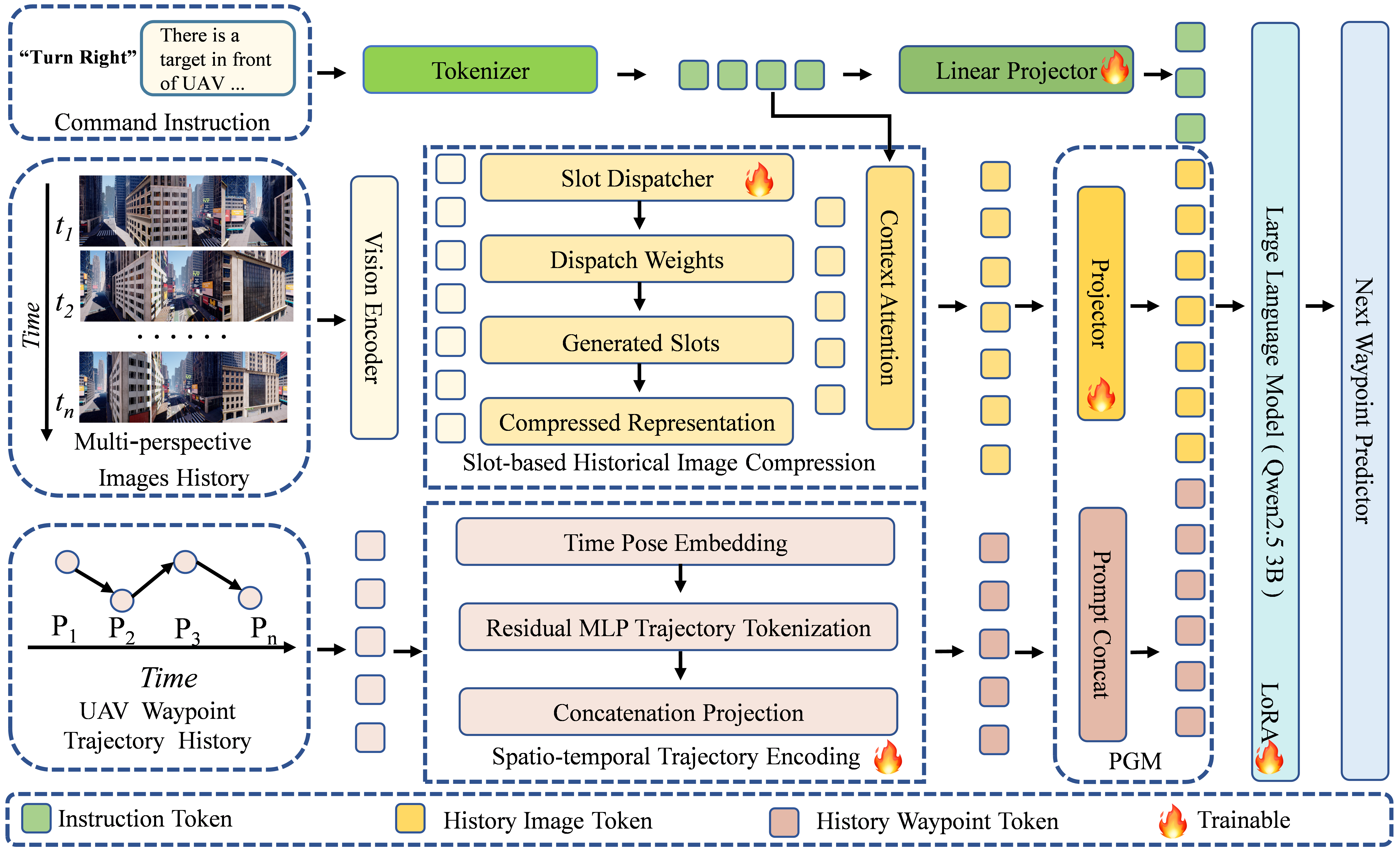}  
    \caption{LongFly is a spatiotemporal context modeling framework for long-horizon UAV VLN. It addresses long-horizon reasoning by jointly integrating language instructions, historical visual observations, and flight trajectories. Given a natural language instruction, the framework maps multi-view historical images and past trajectory points into compact representations. Historical visual observations are compressed to extract instruction-relevant semantic cues, while historical trajectories are encoded as explicit motion priors that capture long-horizon path evolution. These visual and trajectory contexts are then fused under the guidance of the instruction and fed into a multimodal model for cross-modal reasoning, enabling consistent waypoint prediction over long distances.}
    \label{fig:overview}
\end{figure*}

\subsection{Long-horizon vision-and-language navigation for UAVs}

Several recent studies explore long-horizon UAV VLN from different perspectives. Subsequent efforts, such as Citynav~\cite{lee2024citynav} and TravelUAV\cite{wang2024towards}, incorporate large models to better understand the more complex language~\cite{yuan2025nextgenerationllmuavnatural} and environments involved in long-horizon dependencies~\cite{serpiva2025racevla,cai2025flightgpt},\cite{sautenkov2025uav,zhang2025grounded}. Recently, OpenVLN~\cite{lin2025openvln} introduces reinforcement learning to enhance long-horizon navigation planning capabilities. In addition, BEV mapping has been explored better to align history pictures with language instructions~\cite{zhao2025aerialvisionandlanguagenavigationgridbased}. CityAVOS\cite{ji2025autonomousuavvisualobject} uses object-centric 3D semantic map, cognitive map to improve search success and efficiency. Finally, CityNavAgent maintains a city-scale topological memory for long-horizon path reasoning~\cite{zhang2025citynavagentaerialvisionandlanguagenavigation}. SkyVLN~\cite{li2025skyvlnvisionandlanguagenavigationnmpc} introduces a track-back memory for recording and revisiting trajectories; OpenFly~\cite{gao2025openfly} uses a key-frame memory to exploit historical observations. 
These studies indicate that both trajectory history and visual history are essential for effective long-horizon UAV vision-and-language navigation. However, although existing methods attempt to incorporate different forms of memory and historical information, such history is often modeled separately and treated as static cues, without a unified representation aligned with language instructions or the spatiotemporal structure of the navigation process. As a result, in complex and rapidly changing environments, current UAV-VLN models struggle to maintain consistent spatiotemporal context, gradually drifting away from the intended instructions over long-horizon tasks, which leads to inaccurate semantic alignment and unstable path planning. Compared to short-range navigation, this issue becomes more pronounced in long-horizon scenarios, causing a significant degradation in performance. Overall, how to elevate multimodal historical information from static memory to a spatiotemporally grounded context that is tightly coupled with language and navigation remains a key open problem in long-horizon UAV VLN.

\section{METHOD}

\label{sec:method}

\subsection{Problem Formulation}

In UAV VLN, the objective is to guide an UAV through complex 3D environments based on a natural language instruction $L$, which describes the target location and related semantic cues. At each time step $t$, each navigation episode starts with the UAV initialized at a specific position and orientation : \( Q_t = [x_t, y_t, z_t, \varphi_t, \theta_t, \psi_t] \), where $(x_t, y_t, z_t)$ denotes the UAV’s 3D coordinates, and $(\varphi_t, \theta_t, \psi_t)$ represent its pitch, roll, and yaw. What’s more, the UAV captures RGB images $R_t$ from five viewpoints: front, rear, left, right, and bottom, where
$R_t^1$, $R_t^2$, $R_t^3$, $R_t^4$, and $R_t^5$ correspond to the front, rear, left, right, and bottom views, respectively.

Unlike traditional methods that select discrete actions (e.g., forward, turn left), our setting adopts a continuous waypoint prediction paradigm. Specifically, the UAV predicts a 3D waypoint $P_t = [x_t, y_t, z_t]$ at each step, forming a trajectory $\{P_1, P_2, \dots, P_T\}$. The episode terminates when a special ``Stop'' signal is generated or a predefined maximum step count is reached. Navigation is deemed successful if the final predicted waypoint lies within 20 meters of the goal.

To improve long-horizon reasoning, the UAV leverages not only the current observation $(R_t, Q_t)$ and instruction $L$, but also historical information. At the current time step $t$, the UAV leverages historical information spanning from the initial step $t_1$ to the previous step $t_{t-1}$, including historical visual observations and flight trajectories. The historical RGB observation sequence is denoted as $\{R_1, R_2, \dots, R_{t-1}\}$ and is encoded and compressed into a compact visual memory representation using Slot-based historical image compression. The historical waypoint sequence is represented as $\{P_1, P_2, \dots, P_{t-1}\}$, and the corresponding historical UAV pose information is embedded by a trajectory encoder to generate trajectory tokens, capturing the spatiotemporal dynamics of UAV motion.

\subsection{LongFly Overview}
we propose LongFly, a spatiotemporal context modeling framework for long-horizon UAV VLN as shown in Fig.~\ref{fig:overview}. Existing UAV VLN models perform well in short-range tasks but face significant performance degradation in long-horizon tasks. This is due to over-reliance on static historical information retrieval and the lack of spatiotemporal context modeling, reducing robustness. Key challenges include adaptively extracting relevant information from redundant historical data and suppressing noise. Additionally, misalignment between visual observations, flight trajectories, and language instructions hampers the capture of logical relationships between past movements and current goals, leading to inconsistent navigation behavior.

\begin{figure}[htbp]
    \centering
    \includegraphics[width=1\linewidth]{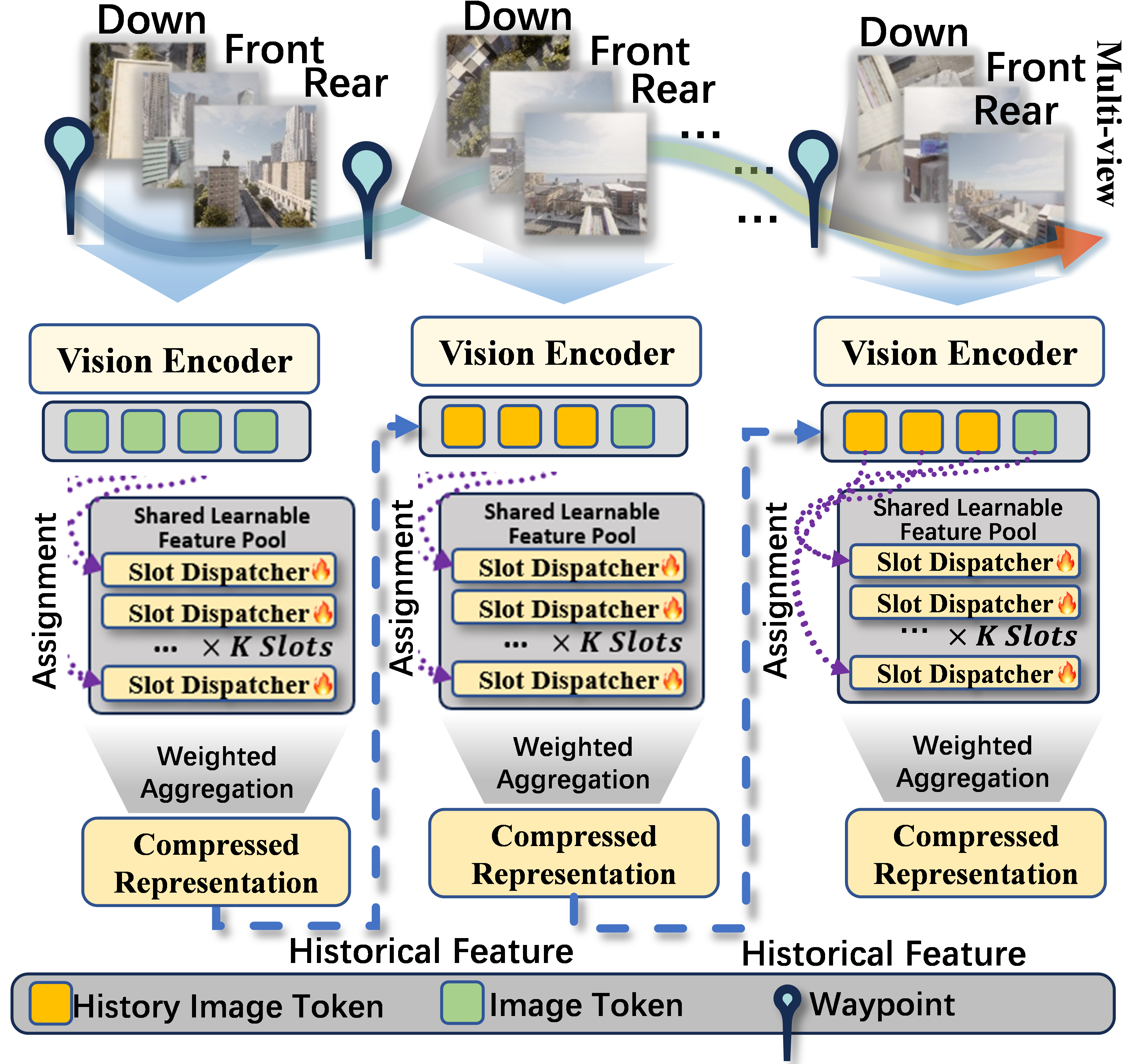}
    \caption{Overview of the Slot-Based Historical Image Compression (SHIC) module.
Multi-view visual observations collected over time are encoded and dynamically aggregated
into a fixed number of semantic slots.
Through recurrent slot assignment and weighted aggregation, SHIC compresses long-horizon
visual histories into compact representations that retain spatial and semantic consistency.}
    \label{fig:shic}
\end{figure}

To address these challenges, LongFly proposes a history-aware spatiotemporal modeling strategy that transforms fragmented and redundant historical data into structured, compact, and expressive representations. Specifically, First, we propose the slot-based historical image compression module, which dynamically distills multi-view historical observations into fixed-length contextual representations. Then, the spatiotemporal trajectory encoding module is introduced to capture the temporal dynamics and spatial structure of UAV trajectories. Finally, to integrate existing spatiotemporal context with current observations, we design the prompt-guided multimodal integration module to support time-based reasoning and robust waypoint prediction. Detailed formulations and implementations of each module are presented in the following subsections.

\subsection{History-Aware Spatiotemporal Context Modeling Strategy}
\subsubsection{Slot-based Historical Image Compression (SHIC)}

In long-horizon UAV VLN, maintaining temporally consistent and semantically rich visual memory is essential for stable decision-making. However, directly storing and processing high-dimensional historical visual features leads to computational costs that grow linearly with time. To address this issue, we propose the slot-based historical image compression (SHIC) module, which compresses historical visual observations into a fixed number of semantic slots through a recurrent slot update mechanism, enabling efficient long-horizon visual context modeling as shown in Fig.~\ref{fig:shic}.

\textbf{Visual Feature Extraction.}
Given the multi-view historical image sequence $\{R_1, R_2, \dots, R_{t-1}\}$,
we extract visual tokens at each time step using a CLIP-based visual encoder $\mathcal{F}_v$:
\begin{equation}
Z_i = \mathcal{F}_v(R_i), \quad i = 1,2,\dots,t-1,
\label{eq:shic_feat}
\end{equation}
where $Z_i = \{z_{i,1}, z_{i,2}, \dots, z_{i,N_i}\}$ denotes the set of visual tokens
extracted from $R_i$, and $z_{i,j} \in \mathbb{R}^{d}$.

\textbf{Recurrent Slot Update.}
SHIC maintains a fixed-capacity set of visual memory slots
\begin{equation}
S_i = \{s_{i,1}, s_{i,2}, \dots, s_{i,K}\}, \quad s_{i,k} \in \mathbb{R}^{d},
\label{eq:shic_slots}
\end{equation}
where $K$ is the number of slots and $d$ is the feature dimension.
At the initial step ($i=1$), the slot set is initialized as learnable parameters:
\begin{equation}
S_1 = \Phi, \quad \Phi \in \mathbb{R}^{K \times d}.
\label{eq:shic_init}
\end{equation}

For each subsequent time step $i = 2,\dots,t-1$, the previous slot set $S_{i-1}$
acts as semantic memory and is updated by interacting with the newly observed tokens $Z_i$.
Specifically, each slot is treated as a query, while visual tokens serve as keys and values:
\begin{equation}
q_{i-1,k} = W_q s_{i-1,k}, \quad
k_{i,j}   = W_k z_{i,j}, \quad
v_{i,j}   = W_v z_{i,j},
\label{eq:shic_qkv}
\end{equation}
where $W_q, W_k, W_v \in \mathbb{R}^{d \times d}$ are learnable projection matrices.

The attention weight between the $k$-th slot and the $j$-th visual token is computed as
\begin{equation}
\alpha_{i,k,j}
=
\frac{
\exp\!\left(\frac{1}{\sqrt{d}} q_{i-1,k}^{\top} k_{i,j}\right)
}{
\sum\limits_{j'=1}^{N_i}
\exp\!\left(\frac{1}{\sqrt{d}} q_{i-1,k}^{\top} k_{i,j'}\right)
},
\label{eq:shic_attn}
\end{equation}
where the softmax normalization is performed over the token dimension $j$ for each slot $k$.

Based on these attention weights, the feature increment for the $k$-th slot at time step $i$ is
\begin{equation}
\hat{s}_{i,k}
=
\sum\limits_{j=1}^{N_i}
\alpha_{i,k,j}\, v_{i,j}.
\label{eq:shic_agg}
\end{equation}

The slot memory is then updated using a gated recurrent unit (GRU):
\begin{equation}
S_i = \mathrm{GRU}(S_{i-1}, \hat{S}_i),
\quad
\hat{S}_i = \{\hat{s}_{i,1}, \dots, \hat{s}_{i,K}\}.
\label{eq:shic_gru}
\end{equation}

\textbf{Compressed Visual Memory.}
After processing the historical image sequence $\{R_1, R_2, \dots, R_{t-1}\}$,
the resulting slot set $S_{t-1}$ serves as a compact visual memory with fixed capacity,
summarizing long-horizon visual observations for downstream reasoning.
Through the recurrent slot update mechanism, SHIC exhibits the following properties. First, the slot representations evolve continuously over time, forming a dynamic semantic memory that captures persistent landmarks and spatial layouts in long-horizon navigation. Second, by compressing variable-length visual histories into a fixed number of slots, the memory and computational complexity during inference is reduced from $O(t)$ to $O(1)$. Finally, independent slot sets are maintained for different camera viewpoints, ensuring that viewpoint-specific semantic cues are preserved during compression.

\begin{figure}[htbp]
    \centering
    \includegraphics[width=1\linewidth]{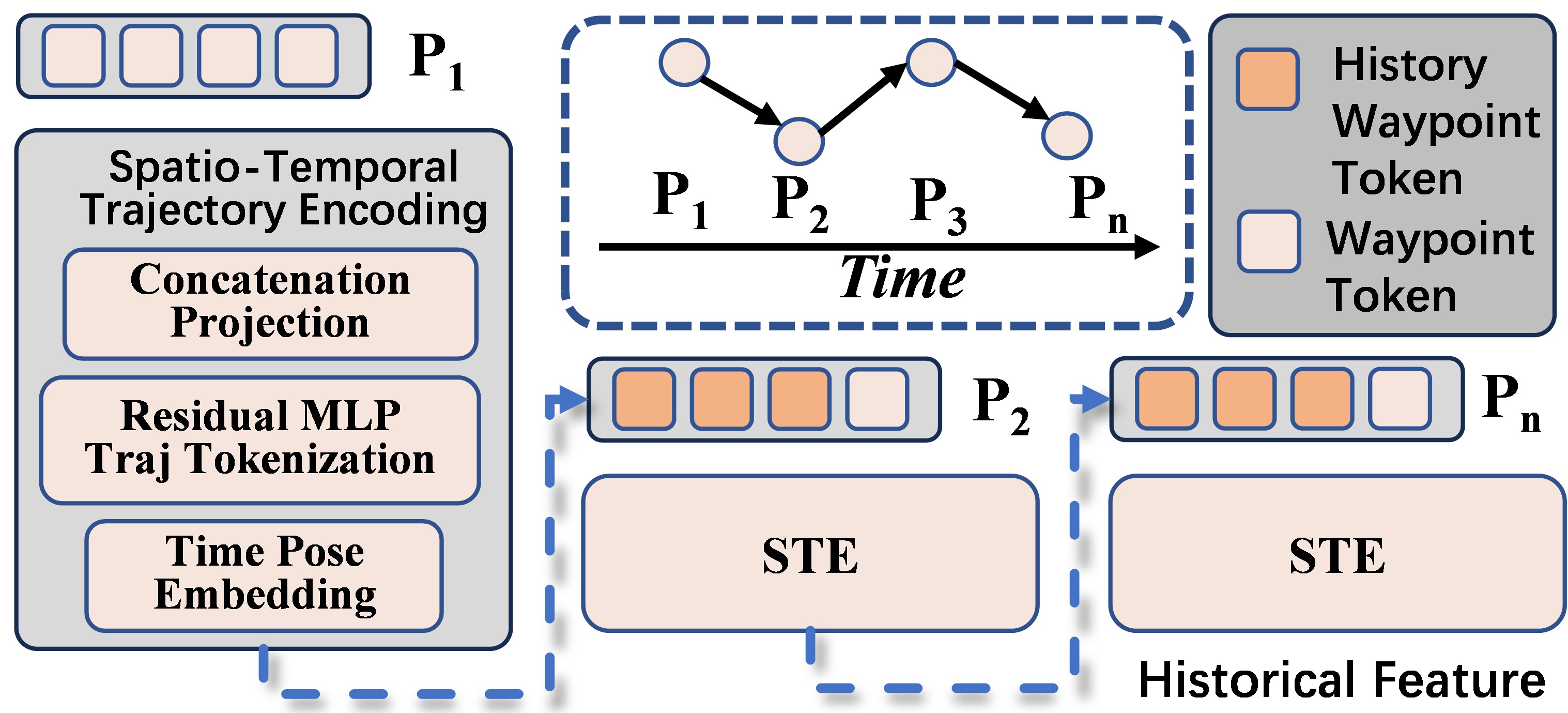}
    \caption{
Structure of the Spatio-Temporal Trajectory Encoding (STE) module.
Historical waypoints are encoded into trajectory tokens in temporal order.
By combining concatenation projection, time–pose embedding, and a residual MLP,
STE models motion continuity and provides spatiotemporal trajectory features
for long-horizon navigation.
}
    \label{fig:ste}
\end{figure}

\subsubsection{Spatio-temporal Trajectory Encoding module (STE)}

To capture the spatiotemporal dynamics of UAV motion, we encode the historical waypoint
trajectory $\{P_1, P_2, \dots, P_{t-1}\}$, where $P_i = [x_i, y_i, z_i] \in \mathbb{R}^3$
denotes the predicted 3D waypoint at time step $i$.
Instead of directly using absolute coordinates, we first transform waypoints into
relative motion representations to reduce sensitivity to global position drift as shown in Fig.~\ref{fig:ste}.

\textbf{Relative Motion Representation.}
For each time step $i>1$, we compute the relative displacement vector:
\begin{equation}
\Delta P_i = P_i - P_{i-1}.
\label{eq:ste_delta}
\end{equation}
The displacement is further decomposed into a unit direction vector and a motion scale:
\begin{equation}
r_i = \lVert \Delta P_i \rVert_2, \quad
\mathbf{d}_i = \frac{\Delta P_i}{\lVert \Delta P_i \rVert_2 + \epsilon},
\label{eq:ste_dir_scale}
\end{equation}
where $r_i \in \mathbb{R}$ denotes the step length, $\mathbf{d}_i \in \mathbb{R}^3$
represents the motion direction, and $\epsilon$ is a small constant for numerical stability.

We then construct a 4D motion descriptor by concatenating direction and scale:
\begin{equation}
M_i = [\, \mathbf{d}_i \, \| \, r_i \,] \in \mathbb{R}^4.
\label{eq:ste_motion}
\end{equation}

\textbf{Temporal Augmentation.}
To encode temporal ordering, each motion descriptor is augmented with a temporal embedding
$\tau_i = \mathcal{E}_t(i)$, where $\mathcal{E}_t(\cdot)$ denotes a sinusoidal or learnable
time embedding function:
\begin{equation}
\widetilde{M}_i = [\, M_i \, \| \, \tau_i \,] \in \mathbb{R}^{4 + d_t}.
\label{eq:ste_time}
\end{equation}

\textbf{Trajectory Token Encoding.}
Each time-aware motion representation $\widetilde{M}_i$ is projected into a
$d$-dimensional trajectory token using an MLP encoder $\mathcal{F}_t(\cdot)$:
\begin{equation}
t_i = \mathcal{F}_t(\widetilde{M}_i)
= \phi\!\left(
W_2\, \phi\!\left(W_1 \widetilde{M}_i + b_1\right) + b_2
\right),
\label{eq:ste_mlp}
\end{equation}
where $W_1, W_2$ and $b_1, b_2$ are learnable parameters, and $\phi(\cdot)$ is a ReLU activation.

Finally, we obtain the trajectory token sequence:
\begin{equation}
T_{t-1} = \{t_2, t_3, \dots, t_{t-1}\},
\label{eq:ste_tokens}
\end{equation}
which serves as an explicit motion prior capturing long-horizon path evolution for downstream
multimodal reasoning.

\begin{figure}[htbp]
    \centering
    \includegraphics[width=1\linewidth]{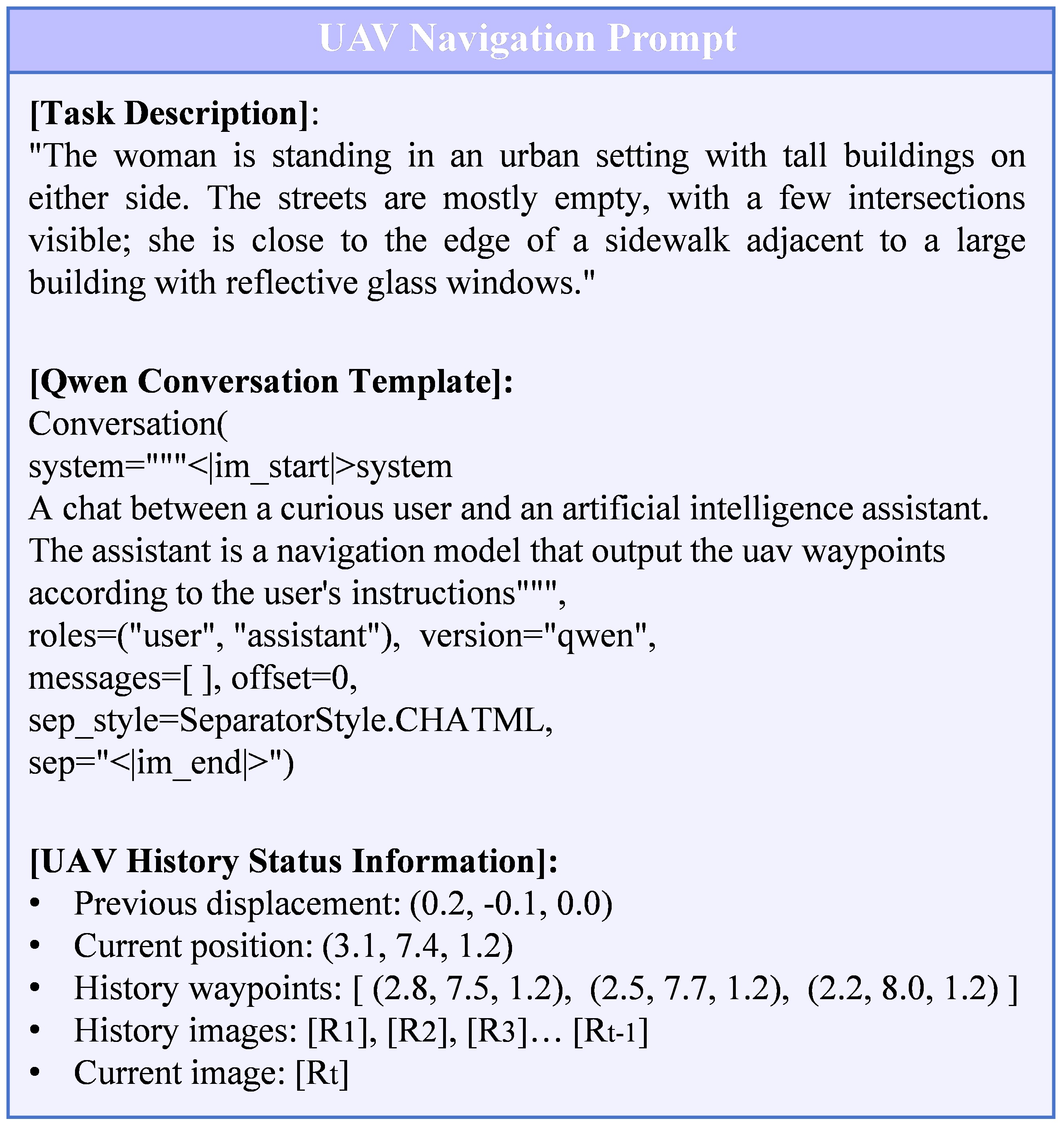}
    \caption{Illustration of the prompt-guided multimodal integration in LongFly.
    The structured prompt consists of three parts: (i) the task instruction,
    (ii) a Qwen-compatible conversation template, and
    (iii) UAV history status information, including previous displacement,
    current position, historical waypoints, historical visual observations,
    and the current image.}
    \label{fig:prompt}
\end{figure}

\subsubsection{Prompt-Guided Multimodal integration module (PGM)}

To integrate historical visual memory and motion history for UAV navigation,
we design a prompt-guided multimodal integration module (PGM).
Instead of introducing additional feature-level fusion mechanisms, PGM organizes multimodal context into a structured prompt and leverages a multimodal large language model to perform instruction-conditioned reasoning and waypoint prediction.

\textbf{Instruction Encoding.}
Given a natural language instruction $L$,
we first encode it using a pretrained BERT encoder $\mathcal{F}_l(\cdot)$
to obtain a contextualized semantic representation:
\begin{equation}
E_L = \mathcal{F}_l(L),
\end{equation}
where $E_L \in \mathbb{R}^{d_l}$ denotes the sentence-level embedding
extracted from the \texttt{[CLS]} token.
To align the instruction representation with the multimodal reasoning space,
we further project it into a unified latent dimension:
\begin{equation}
\widetilde{E}_L = W_L E_L + b_L, \quad \widetilde{E}_L \in \mathbb{R}^{2048},
\end{equation}
where $W_L$ and $b_L$ are learnable projection parameters.

\textbf{Prompt Inputs.}
At time step $t$, PGM collects the following navigation-relevant information:
(i) the projected instruction embedding $\widetilde{E}_L$,
(ii) the compressed historical visual memory from SHIC $S_{t-1}$,
(iii) the encoded historical trajectory tokens from STE $T_{t-1}$,
and (iv) the current visual observation $R_t$.

To ensure consistent interaction with the MLLM,
the historical visual and trajectory representations are projected
into the same latent space:
\begin{equation}
\widetilde{S}_{t-1} = \mathcal{P}_v(S_{t-1}), \quad
\widetilde{T}_{t-1} = \mathcal{P}_m(T_{t-1}),
\end{equation}
where $\mathcal{P}_v(\cdot)$ and $\mathcal{P}_m(\cdot)$ denote linear
projection layers that map visual slots and trajectory tokens
into $\mathbb{R}^{2048}$.

\textbf{Structured Prompt Construction.}
PGM organizes the multimodal context into a structured navigation  as shown in Fig.~\ref{fig:prompt},
explicitly distinguishing different sources of information.
At a semantic level, the prompt at time step $t$ is defined as:
\begin{equation}
\mathcal{P}_t =
\big\{
L,Q,\;
\{P_{1}, \dots, P_{t-1}\},P_{t},\;
\{\mathrm{R}_{1}, \dots, \mathrm{R}_{t-1}\},\;
\mathrm{R}_{t}
\big\},
\end{equation}
where $L$ denotes the task instruction,
$\{P_{1:t-1}\}$ represents the historical waypoint sequence,
and $\{\mathrm{R}_{1:t-1}\}$ together with $\mathrm{R}_t$
denote the historical and current visual observations, respectively.

\textbf{Multimodal Conditioning and Waypoint Prediction.}
The structured prompt $\mathcal{P}_t$ is paired with the current visual input
to form the multimodal input to the MLLM:
\begin{equation}
X_t = \big(\mathcal{P}_t,\ \mathrm{R}_{t}\big).
\end{equation}
The multimodal large language model then predicts the next waypoint in
continuous space:
\begin{equation}
P_{t+1} = \mathrm{LLM}(X_t),
\end{equation}
where $P_{t+1} \in \mathbb{R}^3$ denotes the predicted 3D waypoint.

By serializing language, historical visual memory, and motion priors
into a unified structured prompt aligned with the 2048-dimensional
hidden space of Qwen2.5-3B,
PGM enables coherent long-horizon multimodal reasoning
without introducing additional fusion modules.

\section{EXPERIMENT}

In this section, we first describe the setup (datasets, simulator, metrics, and implementation). We then report quantitative results on benchmark under seen and unseen environments, and further assess generalization on novel objects and unseen maps. We conduct ablations on SHIC, STE, and PGM, and study hyperparameters such as learning rate, history length, and SHIC slot number. Finally, we present qualitative visualizations showing how spatiotemporal context improves multimodal grounding and long-horizon planning.

\subsection{Experimental Setup}
\label{sec:experiment_setup}

\subsubsection{Datasets}

The OpenUAV dataset~\cite{wang2024towards} is designed for UAV object search and navigation tasks, containing 12,149 human-operated flight trajectories with lengths ranging from 50 to 400 meters. Each trajectory is paired with multi-view RGB images (front, back, left, right, top), textual goal descriptions refined by experts, and corresponding waypoint sequences. The dataset covers 89 object categories, including vehicles, humans, and animals, and provides diverse outdoor scenarios for evaluating long-horizon reasoning, multimodal grounding, and generalization in complex, GPS-denied environments.

\subsubsection{Experimental Environment}

\begin{table*}[t]
\caption{Results on the Test Unseen Set across Full/Easy/Hard splits using
NE$\downarrow$  and SR/OSR/SPL(\%)↑;\\
LongFly (ours) is compared with baselines, and absolute gains are reported as 
Imp.(BS) and Imp.(SOTA).}
\label{tab:unseen_results}
\centering
\small
\setlength{\tabcolsep}{4pt}
\begin{adjustbox}{max width=\textwidth} 
\begin{tabular}{lcccccccccccc}
\toprule
\multirow{2}{*}{Method} &
\multicolumn{4}{c}{\textbf{Full}} &
\multicolumn{4}{c}{\textbf{Easy}} &
\multicolumn{4}{c}{\textbf{Hard}} \\
\cmidrule(lr){2-5} \cmidrule(lr){6-9} \cmidrule(lr){10-13}
& NE$\!\downarrow$ & SR(\%)$\!\uparrow$ & OSR(\%)$\!\uparrow$ & SPL(\%)$\!\uparrow$
& NE$\!\downarrow$ & SR(\%)$\!\uparrow$ & OSR(\%)$\!\uparrow$ & SPL(\%)$\!\uparrow$
& NE$\!\downarrow$ & SR(\%)$\!\uparrow$ & OSR(\%)$\!\uparrow$ & SPL(\%)$\!\uparrow$ \\
\midrule
Random Action               & 225.64 & 0.06  & 0.06  & 0.06  & 164.66 & 0.19  & 0.19  & 0.19  & 280.58 & 0.00  & 0.00  & 0.00 \\
Fixed Action                & 193.30 & 1.76  & 5.36  & 1.09  & 140.33 & 3.19  & 8.08  & 1.88  & 245.96 & 0.85  & 3.08  & 0.55 \\
CMA\cite{anderson2018vision}                  & 147.27 & 4.98  & 12.41 & 4.74  & 102.54 & 8.03  & 17.52 & 7.52  & 191.30 & 2.76  & 7.53  & 2.71 \\
TravelUAV\cite{wang2024towards}              & 130.60 & 11.41 & 31.13 & 10.45 & 96.27  & 12.47 & 33.31 & 11.29 & 167.49 & 10.62 & 28.91 & 9.80 \\
NavFoM\cite{zhang2025embodiednavigationfoundationmodel}	& 118.34	& 15.63	& 30.46	& 14.21	& 89.77	& 16.98	& 32.22	& 15.35	& 155.69	& 14.35	& 27.79	& 13.16 \\

\midrule
\rowcolor{blue!10}
BS(Ours)     & 106.08 & 13.99 & 27.66 & 12.16 & 75.25  & 16.73 & 32.40 & 14.01 & 133.49 & 11.55 & 23.45 & 10.52 \\
\rowcolor{blue!20}
\textbf{LongFly(Ours)}                 & \textbf{91.84}  & \textbf{24.19} & \textbf{43.86} & \textbf{20.84}
                     & \textbf{69.16}  & \textbf{22.89} & \textbf{43.24} & \textbf{18.66}
                     & \textbf{112.02} & \textbf{25.36} & \textbf{44.41} & \textbf{22.76} \\
Imp.(BS) & \negv{-14.24} & \pos{+10.20} & \pos{+16.20} & \pos{+8.68}
                 & \negv{-6.09}  & \pos{+6.16} & \pos{+10.84} & \pos{+4.65}
                 & \negv{-21.47} & \pos{+13.81} & \pos{+20.96} & \pos{+12.24} \\

Imp.(SOTA) & \negv{-26.50}  & \pos{+8.56} & \pos{+13.40} & \pos{+6.63}
               & \negv{-20.61}  & \pos{+5.91}  & \pos{+11.02} & \pos{+3.31}
               & \negv{-43.67}  & \pos{+11.01} & \pos{+16.62} & \pos{+9.60} \\

\bottomrule
\end{tabular}
\end{adjustbox}
\end{table*}

The AirSim simulator~\cite{airsim2017fsr} provides a high-fidelity platform for UAV VLN research~\cite{vlmnv2025}, offering diverse urban and natural environments with realistic physics, weather, and lighting conditions. Its flexible object placement and programmable API enable customized scenarios, making it well-suited for developing and evaluating UAV navigation in complex settings.

\begin{figure}[!t]
    \centering  
    \includegraphics[width=0.5\textwidth]{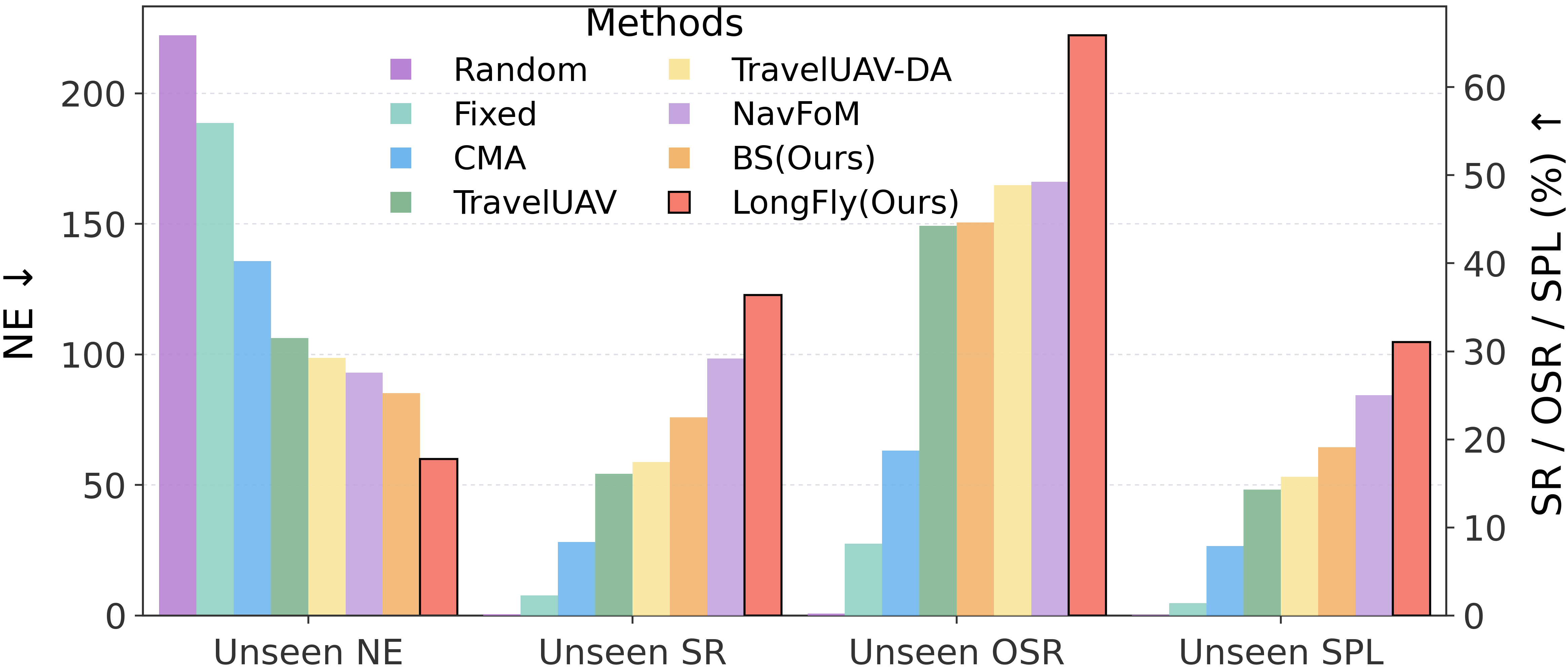}   
    \caption{Evaluation of Navigation Methods Across Test Seen(full) Using SR, OSR, NE, and SPL Metrics.}
    \label{fig:seen6}
\end{figure}

\begin{table*}[t]
\caption{Results on the Test Seen Set across Full/Easy/Hard splits using NE↓  and SR/OSR/SPL(\%)↑; Human upper bound included, with LongFly (ours) compared to baselines and absolute gains reported as Imp.(BS) and Imp.(SOTA). \\BS denotes the Qwen-based base system without SHIC and STE (PGM kept).}
\label{tab:seen_l1_results}
\centering
\small
\setlength{\tabcolsep}{4pt}  
\begin{adjustbox}{max width=\textwidth} 
\begin{tabular}{lcccccccccccc}
\toprule
\multirow{2}{*}{Method} &
\multicolumn{4}{c}{\textbf{Full}} &
\multicolumn{4}{c}{\textbf{Easy}} &
\multicolumn{4}{c}{\textbf{Hard}} \\
\cmidrule(lr){2-5} \cmidrule(lr){6-9} \cmidrule(lr){10-13}
& NE$\!\downarrow$ & SR(\%)$\!\uparrow$ & OSR(\%)$\!\uparrow$ & SPL(\%)$\!\uparrow$
& NE$\!\downarrow$ & SR(\%)$\!\uparrow$ & OSR(\%)$\!\uparrow$ & SPL(\%)$\!\uparrow$
& NE$\!\downarrow$ & SR(\%)$\!\uparrow$ & OSR(\%)$\!\uparrow$ & SPL(\%)$\!\uparrow$ \\
\midrule
\rowcolor{gray!25}
Human      &\textbf{14.15} &\textbf{94.51} &\textbf{94.51} &\textbf{77.84} &\textbf{11.68} &\textbf{95.44} &\textbf{95.44} &\textbf{76.19} &\textbf{17.16} &\textbf{93.37} &\textbf{93.37} &\textbf{79.85} \\
\midrule
Random Action   &222.20 &0.14 &0.21 &0.07 
                &142.07 &0.26 &0.39 &0.13 
                &320.12 &0.00 &0.00 &0.00 \\

Fixed Action    &188.61 &2.27 &8.16 &1.40 
                &121.36 &3.48 &11.48 &2.14 
                &270.69 &0.79 &4.09 &0.49 \\

CMA\cite{anderson2018vision} 
                &135.73 &8.37 &18.72 &7.90 
                &84.89 &11.48 &24.52 &10.68 
                &197.77 &4.57 &11.65 &4.51 \\

TravelUAV\cite{wang2024towards}    & 106.28 & 16.10 & 44.26 & 14.30 & 68.78 & 18.84 & 47.61 & 16.39 & 152.04 & 12.76 & 40.16 & 11.76 \\
TravelUAV-DA\cite{wang2024towards} &  98.66&  17.45 & 48.87 & 15.76 & 66.40 & 20.26 & 51.23 & 18.10 & 138.04& 14.02 & 45.98&  12.90 \\

NavFoM\cite{zhang2025embodiednavigationfoundationmodel}	& 93.05	& 29.17	& 49.24	& 25.03	& 58.98	& 32.91	& 53.16	& 27.87 & 	143.83	& 23.58	& 43.40	& 20.80 \\

\midrule
\rowcolor{blue!10}
BS(Ours)      & 85.17 & 22.50 & 	44.64 &  19.12 & 48.65  &  28.23 & 	54.88 & 23.02 & 127.11 & 	15.91 & 32.88 & 14.64  \\
\rowcolor{blue!20}
\textbf{LongFly(Ours)}  & \textbf{60.02} & \textbf{36.39} & \textbf{65.87} & \textbf{31.07}
         & \textbf{38.10} & \textbf{38.52} & \textbf{71.90} & \textbf{31.24}
         & \textbf{85.20} & \textbf{33.94} & \textbf{58.94} & \textbf{30.88} \\

Imp.(BS) & \negv{-25.15} & \pos{+13.89} & \pos{+21.23} & \pos{+11.95}
                 & \negv{-10.55} & \pos{+10.29} &\pos{+17.02} & \pos{+8.22}
                 & \negv{-41.91} & \pos{+18.03} & \pos{+26.06} & \pos{+16.24} \\
Imp.(SOTA) & \negv{-33.03}  & \pos{+7.22} & \pos{+16.63} & \pos{+6.04}
               & \negv{-20.88}  & \pos{+5.61}  & \pos{+18.74} & \pos{+3.37}
               & \negv{-58.63}  & \pos{+10.36} & \pos{+15.54} & \pos{+10.08} \\  

\bottomrule
\end{tabular}
\end{adjustbox}
\end{table*}

\begin{table*}[t]
\caption{Results on the Test Unseen Object Set across Full/Easy/Hard splits using
NE$\downarrow$ and SR/OSR/SPL(\%)↑; \\
LongFly(ours) is compared with baselines, with absolute gains summarized as 
Imp.(BS) and Imp.(SOTA).}
\label{tab:vln-results-uo}
\centering
\small
\setlength{\tabcolsep}{4pt}
\begin{adjustbox}{max width=\textwidth} 
\begin{tabular}{l*{12}{c}}
\toprule
\multirow[c]{2}{*}{\makecell[c]{Method}} & 
\multicolumn{4}{c}{\textbf{Full}} &
\multicolumn{4}{c}{\textbf{Easy}} &
\multicolumn{4}{c}{\textbf{Hard}} \\
\cmidrule(lr){2-5}\cmidrule(lr){6-9}\cmidrule(lr){10-13}
 & NE$\!\downarrow$ & SR(\%)$\!\uparrow$ & OSR(\%)$\!\uparrow$ & SPL(\%)$\!\uparrow$
 & NE$\!\downarrow$ & SR(\%)$\!\uparrow$ & OSR(\%)$\!\uparrow$ & SPL(\%)$\!\uparrow$
 & NE$\!\downarrow$ & SR(\%)$\!\uparrow$ & OSR(\%)$\!\uparrow$ & SPL(\%)$\!\uparrow$ \\
\midrule
Random Action   & 260.14 & 0.16 & 0.16 & 0.16 & 174.10 & 0.48 & 0.48 & 0.48 & 302.96 & 0.00 & 0.00 & 0.00 \\
Fixed Action    & 212.84 & 3.66 & 9.54  & 2.16 & 151.66 & 6.70 & 13.88 & 3.72 & 243.29 & 2.14 & 7.38 & 1.38 \\
CMA\cite{anderson2018vision}      & 155.79 & 9.06 & 16.06 & 8.68 & 102.92 & 14.83 & 22.49 & 13.90 & 182.09 & 6.19 & 12.86 & 6.08 \\
TravelUAV\cite{wang2024towards}  & {118.11} &{22.42} & {46.90} & {20.51}
         & {86.12}  & {24.40} & {49.28} & {22.03}
         & {134.03} & {21.43} & {45.71} & {19.75} \\
NavFoM\cite{zhang2025embodiednavigationfoundationmodel}	& 108.04	& 29.83	& 47.99	& 27.20	& 70.51	& 32.54	& 50.72	& 29.54	& 133.01	& 28.03	& 46.18	& 25.64 \\

\midrule
\rowcolor{blue!10}
BS(Ours) & 96.13 & 23.84 & 37.83 & 20.96 & 68.06 & 26.37 & 42.28 & 22.16 & 98.84 & 23.57 & 37.25 & 21.49 \\

\rowcolor{blue!20}
\textbf{LongFly(Ours)}     & \textbf{66.74} & \textbf{43.87} & \textbf{64.56} & \textbf{38.39}
         & \textbf{54.84} & \textbf{38.01} & \textbf{56.84} & \textbf{31.36}
         & \textbf{57.07} & \textbf{50.25} & \textbf{74.16} & \textbf{45.27} \\



\bottomrule
\end{tabular}
\end{adjustbox}
\end{table*}

\begin{table*}[t]
\caption{Results on the Test Unseen Map Set across Full/Easy/Hard splits using
NE$\downarrow$ and SR/OSR/SPL(\%)↑;\\ 
LongFly(ours) is compared with baselines, and absolute gains are reported as 
Imp.(BS) and Imp.(SOTA).}
\label{tab:vln-results-um}
\centering
\small
\setlength{\tabcolsep}{4pt}
\begin{adjustbox}{max width=\textwidth} 
\begin{tabular}{l*{12}{c}}
\toprule
\multirow[c]{2}{*}{\makecell[c]{Method}} & 
\multicolumn{4}{c}{\textbf{Full}} &
\multicolumn{4}{c}{\textbf{Easy}} &
\multicolumn{4}{c}{\textbf{Hard}} \\
\cmidrule(lr){2-5}\cmidrule(lr){6-9}\cmidrule(lr){10-13}
 & NE$\!\downarrow$ & SR(\%)$\!\uparrow$ & OSR(\%)$\!\uparrow$ & SPL(\%)$\!\uparrow$
 & NE$\!\downarrow$ & SR(\%)$\!\uparrow$ & OSR(\%)$\!\uparrow$ & SPL(\%)$\!\uparrow$
 & NE$\!\downarrow$ & SR(\%)$\!\uparrow$ & OSR(\%)$\!\uparrow$ & SPL(\%)$\!\uparrow$ \\
\midrule
Random Action   & 202.98 & 0.00 & 0.00 & 0.00 & 158.46 & 0.00 & 0.00 & 0.00 & 265.88 & 0.00 & 0.00 & 0.00 \\
Fixed Action     & 180.47 & 0.52 & 2.61 & 0.39 & 132.89 & 0.89 & 4.28 & 0.67 & 247.72 & 0.00 & 0.25 & 0.00 \\
CMA\cite{anderson2018vision}      & 141.68 & 2.30 & 10.02 & 2.16 & 102.29 & 3.57 & 14.26 & 3.33 & 197.35 & 0.50 & 4.03 & 0.50 \\
TravelUAV\cite{wang2024towards}  & {138.80} & {4.18} & {20.77} & {3.84}
         & {102.94} & {4.63} & {22.82} & {4.24}
         & {189.46} & {3.53} & {17.88} & {3.28} \\
NavFoM\cite{zhang2025embodiednavigationfoundationmodel}	& 125.1	& 6.30	& 18.95	& 5.68	& 102.41	& 6.77	& 20.07	& 6.04	& 170.58	& 5.36	& 15.71	& 4.97 \\

\midrule
\rowcolor{blue!10}
BS(Ours) & 112.61 & 7.52 & 20.98 & 6.38 & 79.97 & 10.40 & 25.91 & 8.66 & 156.24 & 3.66 & 14.39 & 3.32 \\
\rowcolor{blue!20}
\textbf{LongFly(Ours)}     & \textbf{108.32} & \textbf{11.27} & \textbf{30.27} & \textbf{9.32}
         & \textbf{78.56} & \textbf{12.96} & \textbf{34.31} & \textbf{10.32}
         & \textbf{148.10} & \textbf{9.02} & \textbf{24.88} & \textbf{7.98} \\


\bottomrule
\end{tabular}
\end{adjustbox}
\end{table*}

\subsubsection{Evaluation Metrics}

We evaluate navigation performance using four standard metrics widely adopted in VLN tasks: Navigation Error (NE), Success Rate (SR), Oracle Success Rate (OSR) and Success weighted by Path Length (SPL). These metrics were originally introduced for evaluating embodied agents in indoor VLN benchmarks such as R2R\cite{anderson2018vision}, and have since become standard evaluation criteria in both ground-based VLN~\cite{qi2025vlnr1visionlanguagenavigationreinforcement} and UAV VLN\cite{saxena2025uavvlnendtoendvisionlanguage,wu2025aeroduoaerialduouavbased} settings due to their effectiveness in measuring navigation accuracy, efficiency, and robustness.

\subsubsection{Implementation and Training Details}

We adopt a multimodal architecture that integrates both visual and trajectory modalities for UAV navigation. Specifically, for image encoding, we use the CLIP ViT-L/14 model to extract high-level semantic features from RGB observations. Historical images are processed through a slot-based compression mechanism and incorporated into the model via a lightweight prompt design. Historical trajectory points are encoded using a four-layer MLP to capture their temporal dynamics and spatial relevance.

For language processing, we first tokenize the input instructions using a BERT-style tokenizer \cite{devlin2019bert}, ensuring semantic consistency and compatibility with the backbone language model. The language encoding and cross-modal fusion are performed by the pre-trained Qwen-2.5 3B model~\cite{qwen2.5}, which has a hidden size of 2048. All multimodal tokens are projected into a unified embedding space and jointly fused within the Qwen backbone.

\subsection{Training Details}

We conduct our experiments using the PyTorch framework. We train the model on four NVIDIA RTX 4090 GPUs (24 GB each) and perform inference on four NVIDIA A40 GPUs (48 GB each). In this work, BERT-based text encoders and the vision encoder are kept frozen. We update only the SHIC and STE modules. To balance memory and throughput, we use parameter-efficient fine-tuning with LoRA together with ZeRO Stage-2 for distributed optimization. The adapter parameters are trained, while the underlying backbone weights remain fixed. We train the model using a batch size of 8 and an initial learning rate of 5e-4, optimized with AdamW. Scheduled sampling is applied with a decay frequency of 3000 steps and a decay ratio of 0.75, gradually reducing reliance on ground truth supervision to enhance the model’s autonomous prediction ability.

\label{sec:experimental_results}

\subsection{Comparison With State-of-the-Art Methods}

\subsubsection{Quantitative Evaluation on the OpenUAV-Seen Dataset}
we conduct a quantitative evaluation on the OpenUAV Test Seen dataset to assess the performance of various baseline methods:

\begin{itemize}
\item Random Action: The UAV samples waypoint candidates uniformly at random, without planning or guidance.

\item Fixed Action: Instructions are mapped to deterministic macro-actions. 

\item CMA (Cross-Modal Attention)\cite{anderson2018vision}: For our setting, we replace its discrete-action head with a trajectory decoder that outputs waypoint sequences.
\item TravelUAV\cite{wang2024towards}: A UAV VLN baseline on the OpenUAV platform that uses a unified multimodal representation and an action-prediction head for realistic navigation evaluation. 
\item TravelUAV-DA\cite{wang2024towards}: A TravelUAV variant that adds dagger-style data aggregation for closed-loop training, reducing covariate shift and improving trajectory stability.

\item NavFoM (navigation foundation model)\cite{zhang2025embodiednavigationfoundationmodel} is a general-purpose navigation model that takes multi-view video and natural language instructions as input, achieving state-of-the-art or competitive performance on multiple benchmarks without requiring task-specific fine-tuning.

\item BS(Ours): denotes a Qwen-based UAV VLN model without spatiotemporal context Integration, which only leverages the current observation and instruction for waypoint prediction.
\end{itemize}

We present a detailed comparison of different methods on the Test Seen Set under three difficulty levels: Full (the entire test set), Easy (less ambiguous instructions and simpler layouts), and Hard (more complex layouts, longer trajectories, and higher instruction ambiguity). 

As shown in Fig.~\ref{fig:seen6}, LongFly consistently outperforms the baselines in the Full level. LongFly achieves substantial improvements, reducing NE to 60.02 (vs. 98.66 by TravelUAV-DA and 93.05 by NavFoM), while significantly boosting SR (36.39\%) and SPL (31.07\%), demonstrating its ability to generate both accurate and efficient trajectories.


The detailed results are presented in Table~\ref{tab:seen_l1_results}. We compared LongFly against our own baseline and state-of-the-art methods. The results show 33.03m reduction in NE, a 7.22\% gain in SR, and more than 6.04\% improvements in both OSR and SPL. In the Easy subset, although all methods perform relatively better, LongFly remains the top performer. It achieves an SR of 38.52\% and OSR of 71.90\%, indicating its strong capability to follow less ambiguous instructions and leverage simpler environments. Compared to TravelUAV-DA (SR:20.26\%) and NavFoM (SR:32.91\%), LongFly offers a significant performance margin, with over 5.61\% improvement in success rates.

The \textit{Hard} subset poses the greatest challenge. Traditional baselines such as Random and Fixed Action almost entirely fail (SR $\approx$ 0), while stronger models like TravelUAV and NavFoM still suffer from low SR and OSR. In contrast, LongFly achieves a substantial SR of 33.94\%, an OSR of 58.94\%, and a competitive SPL of 30.88\%, demonstrating its robustness in handling complex spatial layouts and semantic ambiguity—especially in scenarios that demand rich multimodal reasoning. LongFly shows its strongest advantage in scenes with complex layouts, long-horizon dependencies, and semantic ambiguity; the concurrent large gains in SR and SPL indicate that it not only reaches the goal more often but also does so along more efficient paths.

Although there remains a sizable gap from the Human upper bound, the unified spatiotemporal context modeling enables LongFly to achieve lower NE and higher SR/SPL across Full, Easy, and Hard levels of the Seen set, with the largest SR/SPL gains on Hard—highlighting robust advantages for long-horizon reasoning and complex semantics. 

\begin{figure*}[!t]
    \centering  
    \includegraphics[width=1\textwidth]{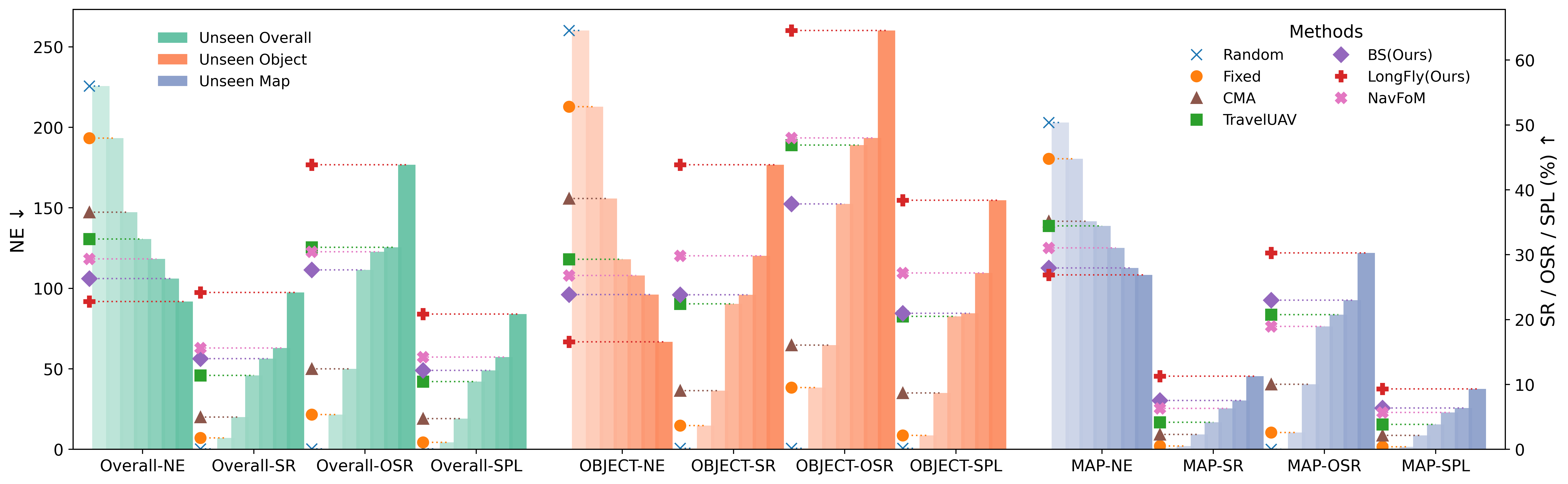}   
    \caption{Evaluation of Navigation Methods Across Unseen Scenarios (Overall, Map, Object) Using NE, SR, OSR and SPL Metrics.}
    \label{fig:grad_full_compact}
\end{figure*}
\subsubsection{Generalization Evaluation on Unseen Environments, Objects, and Maps}

To assess the generalization capability of our proposed LongFly model, we evaluate it under three types of unseen conditions: the overall Test Unseen Set (Table~\ref{tab:unseen_results}), the Test Unseen Object Set (Table~\ref{tab:vln-results-uo}), and the Test Unseen Map Set (Table~\ref{tab:vln-results-um}). We compare against several baselines, including Random, Fixed, CMA, TravelUAV, NavFoM and our own baseline.

\noindent\textbf{Performance on the Overall Test Unseen Set :} On the full Test Unseen set(Table~\ref{tab:unseen_results}), LongFly consistently outperforms all baseline methods across all metrics. For example, on the Full subset, LongFly achieves an SR of 24.19\%, significantly surpassing TravelUAV (11.41\%) and NavFoM (15.63\%). Its OSR reaches 43.86\%, exceeding our baseline (27.66\%) by over 16.20\%. Furthermore, LongFly achieves an SPL of 20.84\%, indicating higher efficiency and path quality. In the Hard subset, LongFly attains 25.36\% SR, 44.41\% OSR, and 22.76\% SPL, while other methods, such as CMA (10.62\%) and NavFoM (14.35\%), fail to effectively handle such challenging conditions. We observe larger overall improvements on long-horizon tasks, indicating that spatiotemporal context modeling is particularly effective in this setting by enabling the model to learn and integrate richer dynamics and spatiotemporal information.

\noindent\textbf{Performance on the Test Unseen Object Set:} The Test Unseen Object Set (Table~\ref{tab:vln-results-uo}) evaluates the model’s ability to generalize to novel object appearances or categories. LongFly achieves an SR of 43.87\% on the Full subset, outperforming NavFoM (29.83\%), and attains an OSR of 64.56\%, clearly surpassing TravelUAV (46.90\%) and NavFoM (47.99\%). On the Easy subset, LongFly reaches 38.01\% SR and 56.84\% OSR. Even in the Hard subset, LongFly maintains robust performance, achieving 74.16\% OSR and 45.27\% SPL, showcasing its capability in both object grounding and path planning. On the Hard split, LongFly shows the largest gains over the SOTA, reducing NE by 75.94m and boosting SR 22.22\% on average.

\noindent\textbf{Performance on the Test Unseen Map Set :} This setting(Table~\ref{tab:vln-results-um}) assesses the model’s ability to adapt to entirely novel scene layouts. On the Full subset, LongFly achieves 11.27\% SR and 30.27\% OSR, significantly outperforming TravelUAV (SR: 4.18\%) and NavFoM (6.30\%). On the Hard subset, LongFly is the only method to maintain reasonable performance, reaching 24.88\% OSR and 7.98\% SPL, while other baselines almost completely fail (OSR $\approx$ 0). In hard scenarios, the NE is reduced 8.14m, but the overall error remains high. Moreover, compared with the unseen-object setting, performance is noticeably worse, indicating that generalization to novel environments is more challenging than to novel objects.

\noindent\textbf{Summary :} As shown in Fig.~\ref{fig:grad_full_compact}, we compare the Full-level results on the three unseen splits—unseen overall, unseen object, and unseen map. LongFly attains the lowest NE and the highest SR/OSR/SPL across all unseen environments. It not only improves navigation performance, but also demonstrates a certain level of generalization, being able to navigate in unseen environments. Taken together, these results underscore that unified spatiotemporal context modeling plays a central role in improving overall performance.
We further observe that LongFly generalizes better to unseen objects than to unseen environments, suggesting that environmental distribution shift is more challenging; therefore, increasing the diversity of training environments is likely to yield larger gains. In addition, on difficult long-horizon tasks, the performance gains are markedly larger than on easy tasks, further demonstrating that spatiotemporal context modeling is especially effective for long-horizon reasoning by enabling the model to learn and integrate richer dynamics and temporal information.

\begin{figure*}[!t]
    \centering
    \includegraphics[width=1\textwidth]{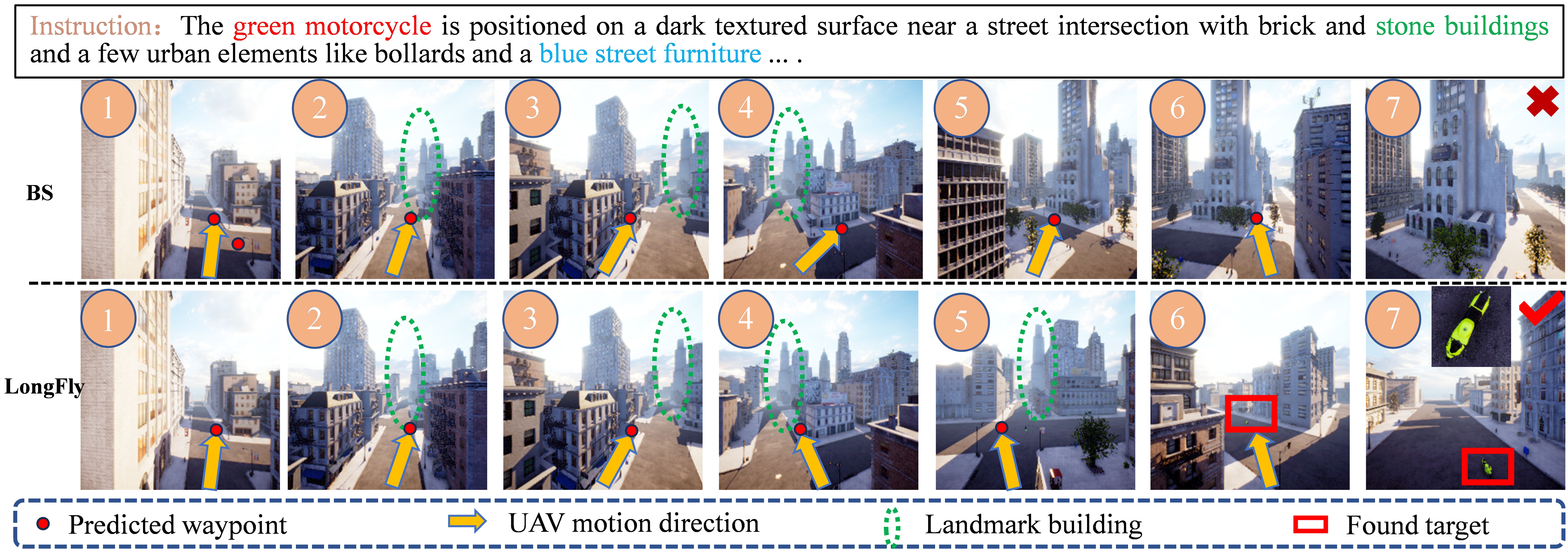}  
    \caption{Qualitative comparison: Top (BS, Qwen-based, no history): the model follows local cues, drifts under viewpoint and layout changes, and fails to reach the target. Bottom (LongFly, a spatiotemporal context modeling framework for long-horizon UAV VLN): by compressing past images and waypoints into slots and aligning them with the instruction, our method maintains consistent localization around landmarks (green dashed ellipses) and reaches the target.}
    \label{fig:oepnview1}
\end{figure*}
\subsection{Ablation Study}
We perform module-wise ablation of LongFly and further assess sensitivity to learning rate, SHIC slot count, and history lengths.

\begin{table*}[t]
\centering
\caption{Ablation of SHIC and STE across Full/Easy/Hard splits across Full/Easy/Hard splits using
NE$\downarrow$ and SR/OSR/SPL(\%)↑; \\ Adding either module improves performance,
and combining both (LongFly = BS+STE+SHIC) is best across all splits.}
\label{tab:ablation_qwen}
\small
\setlength{\tabcolsep}{4pt}
\begin{adjustbox}{max width=\textwidth} 
\begin{tabular}{lcccccccccccc} 
\toprule
\multirow{2}{*}{\textbf{configuration}} &
\multicolumn{4}{c}{\textbf{Full}} &
\multicolumn{4}{c}{\textbf{Easy}} &
\multicolumn{4}{c}{\textbf{Hard}} \\
\cmidrule(lr){2-5} \cmidrule(lr){6-9} \cmidrule(lr){10-13}
 & NE\(\downarrow\) & SR(\%)\(\uparrow\) & OSR(\%)\(\uparrow\) & SPL(\%)\(\uparrow\)
 & NE\(\downarrow\) & SR(\%)\(\uparrow\) & OSR(\%)\(\uparrow\) & SPL(\%)\(\uparrow\)
 & NE\(\downarrow\) & SR(\%)\(\uparrow\) & OSR(\%)\(\uparrow\) & SPL(\%)\(\uparrow\) \\
\midrule
BS & 106.08 & 13.99 & 27.66 & 12.16 &  75.25 & 16.73 & 32.40 & 14.01 & 133.49 & 11.55 & 23.45 & 10.52 \\
BS+STE         & 102.62 & 19.97 & 36.59 & 17.10 &  74.06 & 19.95 & 37.08 & 16.26 & 128.02 & 20.00 & 35.83 & 17.84 \\
BS+SHIC        &  99.24 & 21.05 & 39.70 & 18.15 &  71.34 & 19.28 & 40.96 & 15.84 & 124.05 & 22.62 & 38.57 & 20.20 \\
\rowcolor{blue!10}
BS+STE+SHIC(LongFly)     &  \textbf{91.84} & \textbf{24.19} & \textbf{43.86} & \textbf{20.84} &  \textbf{69.16} & \textbf{22.89} & \textbf{43.24} & \textbf{18.66} & \textbf{112.02} & \textbf{25.36} & \textbf{44.41} & \textbf{22.76} \\
\bottomrule
\end{tabular}
\end{adjustbox}
\end{table*}

\subsubsection{Ablation on Different Components}

To assess the contribution of each module in the LongFly framework, we conduct a series of ablations by progressively adding the SHIC and STE to the baseline. As shown in Table~\ref {tab:ablation_qwen}, adding STE alone (BS+STE) yields consistent improvements over the baseline across all difficulty splits. Adding SHIC alone (BS+SHIC) produces even larger gains, especially in Hard scenarios, highlighting the importance of visual memory for long-horizon navigation. Furthermore, combining both modules (BS+SHIC+STE) achieves the best performance, with a substantial reduction in NE and marked increases in SR and SPL. These results confirm the complementary benefits of temporal history and visual context, and underscore the effectiveness of our prompt-level multimodal fusion strategy in enhancing instruction alignment and robust path planning in complex 3D environments.

\subsubsection{Impact of Learning Rate on Model Performance}

We evaluate three learning rates in the unseen environment: $1\times10^{-4}$, $3\times10^{-4}$, and $5\times10^{-4}$ (Table~\ref{tab:learning_rate}).
Across settings, results are highly stable: the maximum spread is $1.91$ for NE, $0.62$ percentage points for SR, $1.46$ percentage points for OSR, and $0.25$ percentage points for SPL. Among them, $5\times10^{-4}$ achieves the best overall trade-off, with OSR only $0.12$ percentage points below the best value at $3\times10^{-4}$.

\subsubsection{Effect of Prompt Usage and History Length}

We study two factors: (i) whether fusion is guided by prompt tokens and (ii) the length of the visual history. 
First, replacing prompt-guided fusion with plain concatenation of historical cues and the current input leads to a clear drop (Table~\ref{tab:prompt_vs_concat_full}). 
On the \emph{Full} split, SR decreases from 24.19\% to 15.06\% and SPL from 20.84\% to 13.26\%; NE rises from 91.84 to 102.45, which confirms that prompt-based integration is necessary to align historical context with the instruction.

\begin{table}[t]
\centering
\caption{Learning Rate Sensitivity in the Unseen Environment}
\label{tab:learning_rate}
\setlength{\tabcolsep}{4pt}
\begin{tabular}{cccccc}
\toprule
\textbf{ID} & \textbf{Learning Rate} & NE$\!\downarrow$ & SR(\%)$\!\uparrow$ & OSR(\%)$\!\uparrow$ & SPL(\%)$\!\uparrow$ \\
\midrule
1 & $1\times10^{-4}$           & 93.75 & 23.57 & 42.52 & 20.68 \\
2 & $3\times10^{-4}$           & 91.91 & 23.76 & \textbf{43.98} & 20.59 \\
\rowcolor{blue!10}
3 & $5\times10^{-4}$          & \textbf{91.84} & \textbf{24.19} & 43.86 & \textbf{20.84}\\

\bottomrule
\end{tabular}
\end{table}

\begin{table}[t]
\centering
\caption{ Fusion Strategy and History Length (Unseen, Full)}
\label{tab:prompt_vs_concat_full}
\setlength{\tabcolsep}{6pt}
\begin{tabular}{cccccc}
\toprule
\textbf{ID} & \textbf{Strategy} & NE$\!\downarrow$ & SR(\%)$\!\uparrow$ & OSR(\%)$\!\uparrow$ & SPL(\%)$\!\uparrow$ \\
\midrule
1 & LongFly No Prompt          & 102.45 & 15.06 & 31.00 & 13.26 \\
\rowcolor{blue!10}
2 & LongFly       & \textbf{91.84}  & \textbf{24.19} & \textbf{43.86} & \textbf{20.84} \\
\midrule
3 & 10-frame History        & 100.17 & 18.65 & 35.29 & 16.21 \\
4 & 60-frame History        & 97.73 & 20.17 & 38.43 & 17.16 \\
\rowcolor{blue!10}
5 & all-frame History       & \textbf{91.84}  & \textbf{24.19} & \textbf{43.86} & \textbf{20.84} \\
\bottomrule
\end{tabular}
\end{table}

\begin{table}[t]
\centering
\caption{ Effect of SHIC Slot Number $K$ in the Unseen Environment}
\label{tab:slot_k}
\setlength{\tabcolsep}{4pt}
\begin{tabular}{cccccc}
\toprule
\textbf{ID} & \textbf{SHIC Slot $K$} & NE$\!\downarrow$ & SR(\%)$\!\uparrow$ & OSR(\%)$\!\uparrow$ & SPL(\%)$\!\uparrow$ \\
\midrule
1 & 8   & 95.55 & 22.81 & 42.41 & 19.73 \\
2 & 24  & 93.76 & 23.57 & 42.98 & 20.19 \\
\rowcolor{blue!10}
3 & 32  & \textbf{91.84} & \textbf{24.19} & \textbf{43.86} & \textbf{20.84} \\

\bottomrule
\end{tabular}
\end{table}

Second, increasing the number of historical frames consistently improves navigation. 
Moving from 10 to 60 frames raises SR from 18.65\% to 20.17\% and SPL from 16.21\% to 17.16\%, and using all frames yields the best results.

The same trend holds on the \emph{Hard} split, where longer histories bring the largest gains, suggesting that richer temporal context is particularly helpful for long-horizon reasoning, especially on challenging long-horizon tasks.

\subsubsection{Effect of SHIC Slot Number}

We study how the number of SHIC slots, $K$, influences performance by varying $K\in{8,24,32}$ (Table~\ref{tab:slot_k}). As $K$ increases, SR rises from 22.81\% to 24.19\% and SPL from 19.73\% to 20.84\%, while NE decreases from 95.55 to 91.84 and OSR rises from 42.41\% to 43.86\%. We therefore adopt $K=32$ as the default in our main results.

\subsubsection{Qualitative Results}
We compare LongFly with the BS baseline on a language-guided object search task in a complex urban environment, as shown in Fig.~\ref {fig:oepnview1}. Because it lacks spatiotemporal context modeling, BS tends to rely on myopic local cues, making it prone to local traps during long-horizon navigation, gradually deviating from the intended route, and ultimately failing to reach the goal. In contrast, LongFly leverages history-aware spatiotemporal context to fuse past observations and actions with the instruction, maintain global consistency, and select more informative viewpoints.

\section{Conclusion}
\label{sec:conclusions}

In this paper, we present  LongFly, a spatiotemporal context modeling framework for long-horizon UAV VLN. By explicitly leveraging structured historical information, LongFly enables context-aware navigation with better robustness and semantic grounding. On the OpenUAV benchmark, LongFly consistently surpasses prior state-of-the-art methods, improving navigation SR by 7.89\% and SPL by 6.33\% across seen and unseen environments. The gains are largest on long-horizon \textit{Hard} cases, indicating that unified spatiotemporal context modeling is particularly effective for extended reasoning. Ablation studies further confirm that each module contributes, improving both robustness and reasoning ability. For generalization, LongFly performs well in both seen and unseen settings; it generalizes better to unseen objects than to unseen environments, and it remains strong even on never-seen maps.


\bibliographystyle{IEEEtran}
\bibliography{reference}

@inproceedings{anderson2018vision,
  title={Vision-and-language navigation: Interpreting visually-grounded navigation instructions in real environments},
  author={Anderson, Peter and Wu, Qi and Teney, Damien and Bruce, Jake and Johnson, Mark and S{\"u}nderhauf, Niko and Reid, Ian and Gould, Stephen and Van Den Hengel, Anton},
  booktitle={Proceedings of the IEEE Conf. Comput. Vis. Pattern Recognit.},
  pages={3674--3683},
  year={2018}
}

@inproceedings{devlin2019bert,
  title={Bert: Pre-training of deep bidirectional transformers for language understanding},
  author={Devlin, Jacob and Chang, Ming-Wei and Lee, Kenton and Toutanova, Kristina},
  booktitle={Proc. NAACL HLT, volume 1 (long and short papers)},
  pages={4171--4186},
  year={2019}
}

@inproceedings{liu2023aerialvln,
  title={Aerialvln: Vision-and-language navigation for uavs},
  author={Liu, Shubo and Zhang, Hongsheng and Qi, Yuankai and Wang, Peng and Zhang, Yanning and Wu, Qi},
  booktitle={Proceedings of the IEEE/CVF Conf. Comput. Vis. Pattern Recognit.},
  pages={15384--15394},
  year={2023}
}

@article{gao2025openfly,
  title={OpenFly: A Comprehensive Platform for Aerial Vision-Language Navigation},
  author={Gao, Yunpeng and Li, Chenhui and You, Zhongrui and Liu, Junli and Li, Zhen and Chen, Pengan and Chen, Qizhi and Tang, Zhonghan and Wang, Liansheng and Yang, Penghui and others},
  journal={arXiv preprint arXiv:2502.18041},
  year={2025}
}

@inproceedings{airsim2017fsr,
  author = {Shital Shah and Debadeepta Dey and Chris Lovett and Ashish Kapoor},
  title = {AirSim: High-Fidelity Visual and Physical Simulation for Autonomous Vehicles},
  year = {2017},
  booktitle = {Field Serv. Robotics},
  eprint = {arXiv:1705.05065},
  url = {https://arxiv.org/abs/1705.05065}
}

@article{pokhrel2025harnessing,
  title={On Harnessing Semantic Communication With Natural Language Processing},
  author={Pokhrel, Shiva Raj and others},
  journal={IEEE Internet of Things J.},
  year={2025},
  publisher={IEEE}
}

@misc{qwen2.5,
    title = {Qwen2.5: A Party of Foundation Models},
    url = {https://qwenlm.github.io/blog/qwen2.5/},
    author = {Qwen Team},
    month = {September},
    year = {2024}
}

@misc{saxena2025uavvlnendtoendvisionlanguage,
      title={UAV-VLN: End-to-End Vision Language guided Navigation for UAVs}, 
      author={Pranav Saxena and Nishant Raghuvanshi and Neena Goveas},
      year={2025},
      eprint={2504.21432},
      archivePrefix={arXiv},
      primaryClass={cs.RO},
      url={https://arxiv.org/abs/2504.21432}, 
}

@article{Zheng_2024,
   title={ESceme: Vision-and-Language Navigation with Episodic Scene Memory},
   volume={133},
   ISSN={1573-1405},
   url={http://dx.doi.org/10.1007/s11263-024-02159-8},
   DOI={10.1007/s11263-024-02159-8},
   number={1},
   journal={Int. J. Comput. Vis.},
   publisher={Springer Science and Business Media LLC},
   author={Zheng, Qi and Liu, Daqing and Wang, Chaoyue and Zhang, Jing and Wang, Dadong and Tao, Dacheng},
   year={2025},
   month=jul, pages={254–274} }

@ARTICLE{10948476,
  author={Qin, Shaowei and Zhao, Yiji and Wu, Hao and Zhang, Lei and He, Qiang},
  journal={IEEE Trans. Ind. Inform.}, 
  title={Harnessing the Power of Large Language Model for Effective Web API Recommendation}, 
  year={2025},
  volume={21},
  number={7},
  pages={5360-5370},
  doi={10.1109/TII.2025.3552722}}

@misc{wang2023gridmmgridmemorymap,
      title={GridMM: Grid Memory Map for Vision-and-Language Navigation}, 
      author={Zihan Wang and Xiangyang Li and Jiahao Yang and Yeqi Liu and Shuqiang Jiang},
      year={2023},
      eprint={2307.12907},
      archivePrefix={arXiv},
      primaryClass={cs.CV},
      url={https://arxiv.org/abs/2307.12907}, 
}

@article{HE2024110511,
title = {Memory-Adaptive Vision-and-Language Navigation},
journal = {Pattern Recognit.},
volume = {153},
pages = {110511},
year = {2024},
issn = {0031-3203},
doi = {https://doi.org/10.1016/j.patcog.2024.110511},
url = {https://www.sciencedirect.com/science/article/pii/S0031320324002620},
author = {Keji He and Ya Jing and Yan Huang and Zhihe Lu and Dong An and Liang Wang}
}

@misc{li2025skyvlnvisionandlanguagenavigationnmpc,
      title={SkyVLN: Vision-and-Language Navigation and NMPC Control for UAVs in Urban Environments}, 
      author={Tianshun Li and Tianyi Huai and Zhen Li and Yichun Gao and Haoang Li and Xinhu Zheng},
      year={2025},
      eprint={2507.06564},
      archivePrefix={arXiv},
      primaryClass={cs.RO},
      url={https://arxiv.org/abs/2507.06564}, 
}

@misc{zhang2025citynavagentaerialvisionandlanguagenavigation,
      title={CityNavAgent: Aerial Vision-and-Language Navigation with Hierarchical Semantic Planning and Global Memory}, 
      author={Weichen Zhang and Chen Gao and Shiquan Yu and Ruiying Peng and Baining Zhao and Qian Zhang and Jinqiang Cui and Xinlei Chen and Yong Li},
      year={2025},
      eprint={2505.05622},
      archivePrefix={arXiv},
      primaryClass={cs.RO},
      url={https://arxiv.org/abs/2505.05622}, 
}

@misc{xiao2025uavonbenchmarkopenworldobject,
      title={UAV-ON: A Benchmark for Open-World Object Goal Navigation with Aerial Agents}, 
      author={Jianqiang Xiao and Yuexuan Sun and Yixin Shao and Boxi Gan and Rongqiang Liu and Yanjing Wu and Weili Guan and Xiang Deng},
      year={2025},
      eprint={2508.00288},
      archivePrefix={arXiv},
      primaryClass={cs.RO},
      url={https://arxiv.org/abs/2508.00288}, 
}

@misc{zhao2025aerialvisionandlanguagenavigationgridbased,
      title={Aerial Vision-and-Language Navigation with Grid-based View Selection and Map Construction}, 
      author={Ganlong Zhao and Guanbin Li and Jia Pan and Yizhou Yu},
      year={2025},
      eprint={2503.11091},
      archivePrefix={arXiv},
      primaryClass={cs.CV},
      url={https://arxiv.org/abs/2503.11091}, 
}

@misc{ji2025autonomousuavvisualobject,
      title={Towards Autonomous UAV Visual Object Search in City Space: Benchmark and Agentic Methodology}, 
      author={Yatai Ji and Zhengqiu Zhu and Yong Zhao and Beidan Liu and Chen Gao and Yihao Zhao and Sihang Qiu and Yue Hu and Quanjun Yin and Yong Li},
      year={2025},
      eprint={2505.08765},
      archivePrefix={arXiv},
      primaryClass={cs.CV},
      url={https://arxiv.org/abs/2505.08765}, 
}

@misc{zhang2025embodiednavigationfoundationmodel,
      title={Embodied Navigation Foundation Model}, 
      author={Jiazhao Zhang and Anqi Li and Yunpeng Qi and Minghan Li and Jiahang Liu and Shaoan Wang and Haoran Liu and Gengze Zhou and Yuze Wu and Xingxing Li and Yuxin Fan and Wenjun Li and Zhibo Chen and Fei Gao and Qi Wu and Zhizheng Zhang and He Wang},
      year={2025},
      eprint={2509.12129},
      archivePrefix={arXiv},
      primaryClass={cs.RO},
      url={https://arxiv.org/abs/2509.12129}, 
}

@inproceedings{
wang2024towards,
title={Towards Realistic {UAV} Vision-Language Navigation: Platform, Benchmark, and Methodology},
author={Xiangyu Wang and Donglin Yang and Ziqin Wang and Hohin Kwan and Jinyu Chen and Wenjun Wu and Hongsheng Li and Yue Liao and Si Liu},
booktitle={The Thirteenth International Conference on Learning Representations},
year={2025},
url={https://openreview.net/forum?id=rUvCIvI4eB}
}

@ARTICLE{11133703,
  author={Wei, Jiaxing and Sun, Jingwei and Liu, Shaomin and Song, Lisheng and Ma, Yanfei and Xu, Ziwei and Xu, Tongren and Zhou, Ji and Wang, Ziwei and Peng, Zhixing and Wu, Dongxing},
  journal={IEEE Transactions on Geoscience and Remote Sensing}, 
  title={A Robust Framework for Improving Fine-Scale Evapotranspiration Estimation From UAV-Based Multispectral and Thermal Images}, 
  year={2025},
  volume={63},
  number={},
  pages={1-15},
  doi={10.1109/TGRS.2025.3601120}}

@inproceedings{fan2023aerial,
  title={Aerial vision-and-dialog navigation},
  author={Fan, Yue and Chen, Winson and Jiang, Tongzhou and Zhou, Chun and Zhang, Yi and Wang, Xin},
  booktitle={Findings of the Association for Computational Linguistics: ACL 2023},
  pages={3043--3061},
  year={2023}
}

@article{lee2024citynav,
  title={Citynav: Language-goal aerial navigation dataset with geographic information},
  author={Lee, Jungdae and Miyanishi, Taiki and Kurita, Shuhei and Sakamoto, Koya and Azuma, Daichi and Matsuo, Yutaka and Inoue, Nakamasa},
  journal={arXiv preprint arXiv:2406.14240},
  year={2024}
}

@article{cai2025flightgpt,
  title={FlightGPT: Towards Generalizable and Interpretable UAV Vision-and-Language Navigation with Vision-Language Models},
  author={Cai, Hengxing and Dong, Jinhan and Tan, Jingjun and Deng, Jingcheng and Li, Sihang and Gao, Zhifeng and Wang, Haidong and Su, Zicheng and Sumalee, Agachai and Zhong, Renxin},
  journal={arXiv preprint arXiv:2505.12835},
  year={2025}
}

@inproceedings{sautenkov2025uav,
  title={UAV-VLA: Vision-language-action system for large scale aerial mission generation},
  author={Sautenkov, Oleg and Yaqoot, Yasheerah and Lykov, Artem and Mustafa, Muhammad Ahsan and Tadevosyan, Grik and Akhmetkazy, Aibek and Cabrera, Miguel Altamirano and Martynov, Mikhail and Karaf, Sausar and Tsetserukou, Dzmitry},
  booktitle={2025 20th ACM/IEEE International Conference on Human-Robot Interaction (HRI)},
  pages={1588--1592},
  year={2025},
  organization={IEEE}
}

@article{serpiva2025racevla,
  title={Racevla: Vla-based racing drone navigation with human-like behaviour},
  author={Serpiva, Valerii and Lykov, Artem and Myshlyaev, Artyom and Khan, Muhammad Haris and Abdulkarim, Ali Alridha and Sautenkov, Oleg and Tsetserukou, Dzmitry},
  journal={arXiv preprint arXiv:2503.02572},
  year={2025}
}

@article{zhang2025grounded,
  title={Grounded Vision-Language Navigation for UAVs with Open-Vocabulary Goal Understanding},
  author={Zhang, Yuhang and Yu, Haosheng and Xiao, Jiaping and Feroskhan, Mir},
  journal={arXiv preprint arXiv:2506.10756},
  year={2025}
}

@article{lin2025openvln,
  title={OpenVLN: Open-world aerial Vision-Language Navigation},
  author={Lin, Peican and Sun, Gan and Liu, Chenxi and Li, Fazeng and Ren, Weihong and Cong, Yang},
  journal={arXiv preprint arXiv:2511.06182},
  year={2025}
}

@ARTICLE{10476501,
  author={Wang, Yufeng and Fang, Shuangkang and Zhang, Huayu and Li, Hongguang and Zhang, Zehao and Zeng, Xianlin and Ding, Wenrui},
  journal={IEEE Transactions on Geoscience and Remote Sensing}, 
  title={UAV-ENeRF: Text-Driven UAV Scene Editing With Neural Radiance Fields}, 
  year={2024},
  volume={62},
  number={},
  pages={1-14},
  doi={10.1109/TGRS.2024.3379649}}

@INPROCEEDINGS{10640736,
  author={Guo, Haonan and Su, Xin and Wu, Chen and Du, Bo and Zhang, Liangpei and Li, Deren},
  booktitle={IGARSS 2024 - 2024 IEEE International Geoscience and Remote Sensing Symposium}, 
  title={Remote Sensing ChatGPT: Solving Remote Sensing Tasks with ChatGPT and Visual Models}, 
  year={2024},
  volume={},
  number={},
  pages={11474-11478},
  doi={10.1109/IGARSS53475.2024.10640736}}

@ARTICLE{10048552,
  author={Collins, Adam M. and O’Dea, Annika and Brodie, Katherine L. and Bak, A. Spicer and Hesser, Tyler J. and Spore, Nicholas J. and Farthing, Matthew W.},
  journal={IEEE Transactions on Geoscience and Remote Sensing}, 
  title={Automated Extraction of a Depth-Defined Wave Runup Time Series From Lidar Data Using Deep Learning}, 
  year={2023},
  volume={61},
  number={},
  pages={1-13},
  doi={10.1109/TGRS.2023.3244488}}

@ARTICLE{10570242,
  author={Sung, Taejun and Kang, Yoojin and Im, Jungho},
  journal={IEEE Transactions on Geoscience and Remote Sensing}, 
  title={Enhancing Satellite-Based Wildfire Monitoring: Advanced Contextual Model Using Environmental and Structural Information}, 
  year={2024},
  volume={62},
  number={},
  pages={1-16},
  doi={10.1109/TGRS.2024.3418475}}

@ARTICLE{10356107,
  author={Chen, Yuan and Jiang, Jie},
  journal={IEEE Transactions on Geoscience and Remote Sensing}, 
  title={An Oblique-Robust Absolute Visual Localization Method for GPS-Denied UAV With Satellite Imagery}, 
  year={2024},
  volume={62},
  number={},
  pages={1-13},
  doi={10.1109/TGRS.2023.3342142}}

@misc{brown2020languagemodelsfewshotlearners,
      title={Language Models are Few-Shot Learners}, 
      author={Tom B. Brown and Benjamin Mann and Nick Ryder and Melanie Subbiah and Jared Kaplan and Prafulla Dhariwal and Arvind Neelakantan and Pranav Shyam and Girish Sastry and Amanda Askell and Sandhini Agarwal and Ariel Herbert-Voss and Gretchen Krueger and Tom Henighan and Rewon Child and Aditya Ramesh and Daniel M. Ziegler and Jeffrey Wu and Clemens Winter and Christopher Hesse and Mark Chen and Eric Sigler and Mateusz Litwin and Scott Gray and Benjamin Chess and Jack Clark and Christopher Berner and Sam McCandlish and Alec Radford and Ilya Sutskever and Dario Amodei},
      year={2020},
      eprint={2005.14165},
      archivePrefix={arXiv},
      primaryClass={cs.CL},
      url={https://arxiv.org/abs/2005.14165}, 
}

@misc{qi2025vlnr1visionlanguagenavigationreinforcement,
      title={VLN-R1: Vision-Language Navigation via Reinforcement Fine-Tuning}, 
      author={Zhangyang Qi and Zhixiong Zhang and Yizhou Yu and Jiaqi Wang and Hengshuang Zhao},
      year={2025},
      eprint={2506.17221},
      archivePrefix={arXiv},
      primaryClass={cs.CV},
      url={https://arxiv.org/abs/2506.17221}, 
}

@misc{wu2025aeroduoaerialduouavbased,
      title={AeroDuo: Aerial Duo for UAV-based Vision and Language Navigation}, 
      author={Ruipu Wu and Yige Zhang and Jinyu Chen and Linjiang Huang and Shifeng Zhang and Xu Zhou and Liang Wang and Si Liu},
      year={2025},
      eprint={2508.15232},
      archivePrefix={arXiv},
      primaryClass={cs.CV},
      url={https://arxiv.org/abs/2508.15232}, 
}

@misc{zhang2025groundedvisionlanguagenavigationuavs,
      title={Grounded Vision-Language Navigation for UAVs with Open-Vocabulary Goal Understanding}, 
      author={Yuhang Zhang and Haosheng Yu and Jiaping Xiao and Mir Feroskhan},
      year={2025},
      eprint={2506.10756},
      archivePrefix={arXiv},
      primaryClass={cs.RO},
      url={https://arxiv.org/abs/2506.10756}, 
}

@misc{song2025longhorizonvisionlanguagenavigationplatform,
      title={Towards Long-Horizon Vision-Language Navigation: Platform, Benchmark and Method}, 
      author={Xinshuai Song and Weixing Chen and Yang Liu and Weikai Chen and Guanbin Li and Liang Lin},
      year={2025},
      eprint={2412.09082},
      archivePrefix={arXiv},
      primaryClass={cs.CV},
      url={https://arxiv.org/abs/2412.09082}, 
}

@article{LV2025110075,
title = {Hierarchical reinforcement learning method for long-horizon path planning of stratospheric airship},
journal = {Aerospace Science and Technology},
volume = {160},
pages = {110075},
year = {2025},
issn = {1270-9638},
doi = {https://doi.org/10.1016/j.ast.2025.110075},
url = {https://www.sciencedirect.com/science/article/pii/S1270963825001464},
author = {Chao Lv and Ming Zhu and Xiao Guo and Jiajun Ou and Wenjie Lou}
}

@misc{liu2023visualinstructiontuning,
      title={Visual Instruction Tuning}, 
      author={Haotian Liu and Chunyuan Li and Qingyang Wu and Yong Jae Lee},
      year={2023},
      eprint={2304.08485},
      archivePrefix={arXiv},
      primaryClass={cs.CV},
      url={https://arxiv.org/abs/2304.08485}, 
}

@misc{zheng2023judgingllmasajudgemtbenchchatbot,
      title={Judging LLM-as-a-Judge with MT-Bench and Chatbot Arena}, 
      author={Lianmin Zheng and Wei-Lin Chiang and Ying Sheng and Siyuan Zhuang and Zhanghao Wu and Yonghao Zhuang and Zi Lin and Zhuohan Li and Dacheng Li and Eric P. Xing and Hao Zhang and Joseph E. Gonzalez and Ion Stoica},
      year={2023},
      eprint={2306.05685},
      archivePrefix={arXiv},
      primaryClass={cs.CL},
      url={https://arxiv.org/abs/2306.05685}, 
}

@misc{fang2025taskorientedcommunicationsvisualnavigation,
      title={Task-Oriented Communications for Visual Navigation with Edge-Aerial Collaboration in Low Altitude Economy}, 
      author={Zhengru Fang and Zhenghao Liu and Jingjing Wang and Senkang Hu and Yu Guo and Yiqin Deng and Yuguang Fang},
      year={2025},
      eprint={2504.18317},
      archivePrefix={arXiv},
      primaryClass={cs.CV},
      url={https://arxiv.org/abs/2504.18317}, 
}

@misc{zhang2025coordfieldcoordinationfieldagentic,
      title={CoordField: Coordination Field for Agentic UAV Task Allocation In Low-altitude Urban Scenarios}, 
      author={Tengchao Zhang and Yonglin Tian and Fei Lin and Jun Huang and Patrik P. Süli and Qinghua Ni and Rui Qin and Xiao Wang and Fei-Yue Wang},
      year={2025},
      eprint={2505.00091},
      archivePrefix={arXiv},
      primaryClass={cs.RO},
      url={https://arxiv.org/abs/2505.00091}, 
}

@misc{guo2025bedicomprehensivebenchmarkevaluating,
      title={BEDI: A Comprehensive Benchmark for Evaluating Embodied Agents on UAVs}, 
      author={Mingning Guo and Mengwei Wu and Jiarun He and Shaoxian Li and Haifeng Li and Chao Tao},
      year={2025},
      eprint={2505.18229},
      archivePrefix={arXiv},
      primaryClass={cs.RO},
      url={https://arxiv.org/abs/2505.18229}, 
}

@misc{pardyl2025flysearchexploringvisionlanguagemodels,
      title={FlySearch: Exploring how vision-language models explore}, 
      author={Adam Pardyl and Dominik Matuszek and Mateusz Przebieracz and Marek Cygan and Bartosz Zieliński and Maciej Wołczyk},
      year={2025},
      eprint={2506.02896},
      archivePrefix={arXiv},
      primaryClass={cs.CV},
      url={https://arxiv.org/abs/2506.02896}, 
}

@misc{cai2025sagcssemanticawaregaussiancurriculum,
      title={SA-GCS: Semantic-Aware Gaussian Curriculum Scheduling for UAV Vision-Language Navigation}, 
      author={Hengxing Cai and Jinhan Dong and Yijie Rao and Jingcheng Deng and Jingjun Tan and Qien Chen and Haidong Wang and Zhen Wang and Shiyu Huang and Agachai Sumalee and Renxin Zhong},
      year={2025},
      eprint={2508.00390},
      archivePrefix={arXiv},
      primaryClass={cs.CL},
      url={https://arxiv.org/abs/2508.00390}, 
}

@misc{hu2025seepointflylearningfree,
      title={See, Point, Fly: A Learning-Free VLM Framework for Universal Unmanned Aerial Navigation}, 
      author={Chih Yao Hu and Yang-Sen Lin and Yuna Lee and Chih-Hai Su and Jie-Ying Lee and Shr-Ruei Tsai and Chin-Yang Lin and Kuan-Wen Chen and Tsung-Wei Ke and Yu-Lun Liu},
      year={2025},
      eprint={2509.22653},
      archivePrefix={arXiv},
      primaryClass={cs.RO},
      url={https://arxiv.org/abs/2509.22653}, 
}

@misc{chen2025gradnavvisionlanguagemodelenabled,
      title={GRaD-Nav++: Vision-Language Model Enabled Visual Drone Navigation with Gaussian Radiance Fields and Differentiable Dynamics}, 
      author={Qianzhong Chen and Naixiang Gao and Suning Huang and JunEn Low and Timothy Chen and Jiankai Sun and Mac Schwager},
      year={2025},
      eprint={2506.14009},
      archivePrefix={arXiv},
      primaryClass={cs.RO},
      url={https://arxiv.org/abs/2506.14009}, 
}

@misc{verraest2025skydreamerinterpretableendtoendvisionbased,
      title={SkyDreamer: Interpretable End-to-End Vision-Based Drone Racing with Model-Based Reinforcement Learning}, 
      author={Aderik Verraest and Stavrow Bahnam and Robin Ferede and Guido de Croon and Christophe De Wagter},
      year={2025},
      eprint={2510.14783},
      archivePrefix={arXiv},
      primaryClass={cs.RO},
      url={https://arxiv.org/abs/2510.14783}, 
}

@misc{yuan2025nextgenerationllmuavnatural,
      title={Next-Generation LLM for UAV: From Natural Language to Autonomous Flight}, 
      author={Liangqi Yuan and Chuhao Deng and Dong-Jun Han and Inseok Hwang and Sabine Brunswicker and Christopher G. Brinton},
      year={2025},
      eprint={2510.21739},
      archivePrefix={arXiv},
      primaryClass={cs.RO},
      url={https://arxiv.org/abs/2510.21739}, 
}

@misc{ferrag2025uavbenchopenbenchmarkdataset,
      title={UAVBench: An Open Benchmark Dataset for Autonomous and Agentic AI UAV Systems via LLM-Generated Flight Scenarios}, 
      author={Mohamed Amine Ferrag and Abderrahmane Lakas and Merouane Debbah},
      year={2025},
      eprint={2511.11252},
      archivePrefix={arXiv},
      primaryClass={cs.AI},
      url={https://arxiv.org/abs/2511.11252}, 
}

@misc{xiang2025navr2dualrelationreasoninggeneralizable,
      title={Nav-$R^2$ Dual-Relation Reasoning for Generalizable Open-Vocabulary Object-Goal Navigation}, 
      author={Wentao Xiang and Haokang Zhang and Tianhang Yang and Zedong Chu and Ruihang Chu and Shichao Xie and Yujian Yuan and Jian Sun and Zhining Gu and Junjie Wang and Xiaolong Wu and Mu Xu and Yujiu Yang},
      year={2025},
      eprint={2512.02400},
      archivePrefix={arXiv},
      primaryClass={cs.CV},
      url={https://arxiv.org/abs/2512.02400}, 
}

@misc{wu2025vlaanefficientonboardvisionlanguageaction,
      title={VLA-AN: An Efficient and Onboard Vision-Language-Action Framework for Aerial Navigation in Complex Environments}, 
      author={Yuze Wu and Mo Zhu and Xingxing Li and Yuheng Du and Yuxin Fan and Wenjun Li and Zhichao Han and Xin Zhou and Fei Gao},
      year={2025},
      eprint={2512.15258},
      archivePrefix={arXiv},
      primaryClass={cs.RO},
      url={https://arxiv.org/abs/2512.15258}, 
}


\end{document}